\documentclass[journal]{IEEEtran}
\IEEEoverridecommandlockouts
\usepackage{color}
\usepackage{tikz}
\usetikzlibrary{shapes,arrows,arrows.meta}
\usepackage{mathtools}
\usepackage{amssymb}
\usepackage{lipsum}
\usepackage{float}
\usepackage{makecell}
\usepackage{longtable}
\usepackage{stfloats}
\usepackage{multirow}
\usepackage{amsmath}
\usepackage{bm}
\usepackage{algorithm}
\usepackage{algpseudocode}
\usepackage{booktabs}
\usepackage[numbers,sort&compress]{natbib}
\usepackage{balance}
\usepackage{soul}
\algdef{SE}[SUBALG]{Indent}{EndIndent}{}{\algorithmicend\ }%
\algtext*{Indent}
\algtext*{EndIndent}
\usepackage{xcolor}
\usepackage{threeparttable}
\usepackage{natbib}
\usepackage[breaklinks,citecolor=red,urlcolor=red,bookmarks=false,hypertexnames=true]{hyperref}
\allowdisplaybreaks
\usepackage{graphicx}
\usepackage{caption}
\usepackage{subfigure}
\usepackage{todonotes}
\usepackage{comment}
\makeatletter

\newcommand{\Rmnum}[1]{\expandafter\@slowromancap\romannumeral #1@}
\makeatother

\begin{document}

\title{SMURF: Spatial Multi-Representation Fusion for 3D Object Detection with 4D Imaging Radar}

\author{\IEEEauthorblockA{
Jianan Liu\IEEEauthorrefmark{1},
Qiuchi Zhao\IEEEauthorrefmark{1},
Weiyi Xiong,
Tao Huang,~\IEEEmembership{Senior Member,~IEEE,}\\
Qing-Long Han,~\IEEEmembership{Fellow,~IEEE,} 
and Bing Zhu\IEEEauthorrefmark{2},~\IEEEmembership{Member,~IEEE}
}
\vspace{-5 mm}

\thanks{This work has been submitted to the IEEE for possible publication. Copyright may be transferred without notice, after which this version may no longer be accessible.}

%\thanks{The work of Q.~Zhao, W.~Xiong and B.~Zhu was supported by National Natural Science Foundation of China under grant 62073015, and the work of T.~Huang was supported by the Australian Government through the Australian Research Council’s Discovery Projects Funding Scheme under Project DP220101634.}

\thanks{Q.~Zhao, W.~Xiong and B.~Zhu are with the School of Automation Science and Electrical Engineering, Beihang University, Beijing 100191, P.R.~China. Email:
qiuchizhao@buaa.edu.cn (Q. Zhao);
weiyixiong@buaa.edu.cn (W. Xiong);
zhubing@buaa.edu.cn (B. Zhu).}
\thanks{J.~Liu is with Vitalent Consulting, Gothenburg, Sweden. Email: jianan.liu@vitalent.se.}
\thanks{T.~Huang is with the College of Science and Engineering, James Cook University, Cairns QLD 4878, Australia. Email: tao.huang1@jcu.edu.au.}
\thanks{Q.-L.~Han is with the School of Science, Computing and Engineering Technologies, Swinburne University of Technology, Melbourne, VIC 3122, Australia. Email: qhan@swin.edu.au.}
\thanks{\IEEEauthorrefmark{1}Both authors contribute equally to the work and are co-first authors.}
\thanks{\IEEEauthorrefmark{2}Corresponding author.}
}

\maketitle

\begin{abstract}
The 4D millimeter-Wave (mmWave) radar is a promising technology for vehicle sensing due to its cost-effectiveness and operability in adverse weather conditions. However, the adoption of this technology has been hindered by sparsity and noise issues in radar point cloud data. This paper introduces spatial multi-representation fusion (SMURF), a novel approach to 3D object detection using a single 4D imaging radar. SMURF leverages multiple representations of radar detection points, including pillarization and density features of a multi-dimensional Gaussian mixture distribution through kernel density estimation (KDE). KDE effectively mitigates measurement inaccuracy caused by limited angular resolution and multi-path propagation of radar signals. Additionally, KDE helps alleviate point cloud sparsity by capturing density features.
Experimental evaluations on View-of-Delft (VoD) and TJ4DRadSet datasets demonstrate the effectiveness and generalization ability of SMURF, outperforming recently proposed 4D imaging radar-based single-representation models. Moreover, while using 4D imaging radar only, SMURF still achieves comparable performance to the state-of-the-art 4D imaging radar and camera fusion-based method, with an increase of 1.22$\%$ in the mean average precision on bird's-eye view of TJ4DRadSet dataset and 1.32$\%$ in the 3D mean average precision on the entire annotated area of VoD dataset. Our proposed method demonstrates impressive inference time and addresses the challenges of real-time detection, with the inference time no more than 0.05 seconds for most scans on both datasets. This research highlights the benefits of 4D mmWave radar and is a strong benchmark for subsequent works regarding 3D object detection with 4D imaging radar.
\end{abstract}


\IEEEpeerreviewmaketitle
\setcitestyle{square}

\section{Introduction}
\label{introduction}

\IEEEPARstart{C}ONVENTIONAL automotive radar has been extensively utilized in advanced driver assistance systems (ADAS) and autonomous driving \cite{automotive_radar_survey}, with potential applications in future cooperative perception systems \cite{Sun2023Vision}. However, compared to LiDAR-based perception \cite{CenterPoint}\cite{LiDAR_based_MOT_GNNPMB}%\cite{LEGO}
, conventional radar-based perception technologies \cite{ordinary_radar_based_3d_bev_object_detection_point_processing_2}\cite{ordinary_radar_based_3d_bev_object_detection_graph_1}\cite{camera_radar_det_level_fusion_tracking} often encounter limitations such as the absence of elevation information and low resolution. These limitations impede their ability to detect and localize objects in the surrounding environment accurately. In recent years, the development of 4D imaging radar \cite{4d_imaging_radar_survey} emerged as a promising solution to overcome these limitations. 4D imaging radar can measure the pitch angle, enhancing the understanding of the environment and improving object detection and localization accuracy. The evolution of 4D radar and its performance have been explored in \cite{intro_example_4d_radar_1}\cite{intro_example_4d_radar_2}\cite{intro_example_4d_radar_3}. 

The signal processing scheme for 4D radar typically involves the utilization of multiple-input and multiple-output (MIMO) arrays for data acquisition, range-Doppler (RD) coherent processing, direction-of-arrival (DOA) estimation, and point cloud generation, as depicted in \cite{intro_example_4d_radar_4}\cite{intro_example_4d_radar_5}\cite{4d_imaging_radar_point_cloud_generation_0}. Further studies focused on point cloud generation for 4D radar have also emerged. For example, the RPDNet \cite{4d_imaging_radar_point_cloud_generation_1} based on UNet \cite{unet} is proposed for radar point detection, and position coding is introduced in \cite{4d_imaging_radar_point_cloud_generation_2} 
to enhance spatial information extracted from the RD map. Furthermore, \cite{4d_imaging_radar_point_cloud_generation_addition_1} capitalizes on the sparsity of the spectrum and antenna arrays to reduce the cost in point cloud generation. These studies have significantly improved 4D radar's accuracy and density in applications such as ADAS and autonomous driving.

LiDAR point cloud and 4D radar point cloud exhibit several similarities. However, the measurements obtained from 4D radar are subject to noise, primarily stemming from the multi-path propagation of radar signals. 
Additionally, 4D radar captures less geometry and semantic information than the more dense LiDAR point cloud. As a result, existing 3D object detection algorithms specifically developed for dense LiDAR point cloud may yield suboptimal performance when directly applied to sparse 4D radar point cloud data. Meanwhile, some existing models proposed for conventional automotive radar-based object detection might be directly extended for 4D radar-based object detection. For instance, researchers have proposed PointNet-based \cite{ordinary_radar_based_3d_bev_object_detection_point_processing_2}\cite{ordinary_radar_based_3d_bev_object_detection_point_processing_1}\cite{ordinary_radar_based_3d_bev_object_detection_point_processing_3} or graph-convolution-based \cite{ordinary_radar_based_3d_bev_object_detection_graph_1}\cite{ordinary_radar_based_3d_bev_object_detection_graph_2}\cite{ordinary_radar_based_3d_bev_object_detection_graph_3} methods as potential solutions. These end-to-end learning methods offer advantages such as larger receptive fields and simplified structures. However,  
conventional radar systems suffer from limitations in measuring the height of objects. As a result, directly applying these conventional radar-based algorithms to 4D radar is unsuitable.

\begin{figure}[tbp]
\centering
\includegraphics[width=\linewidth]{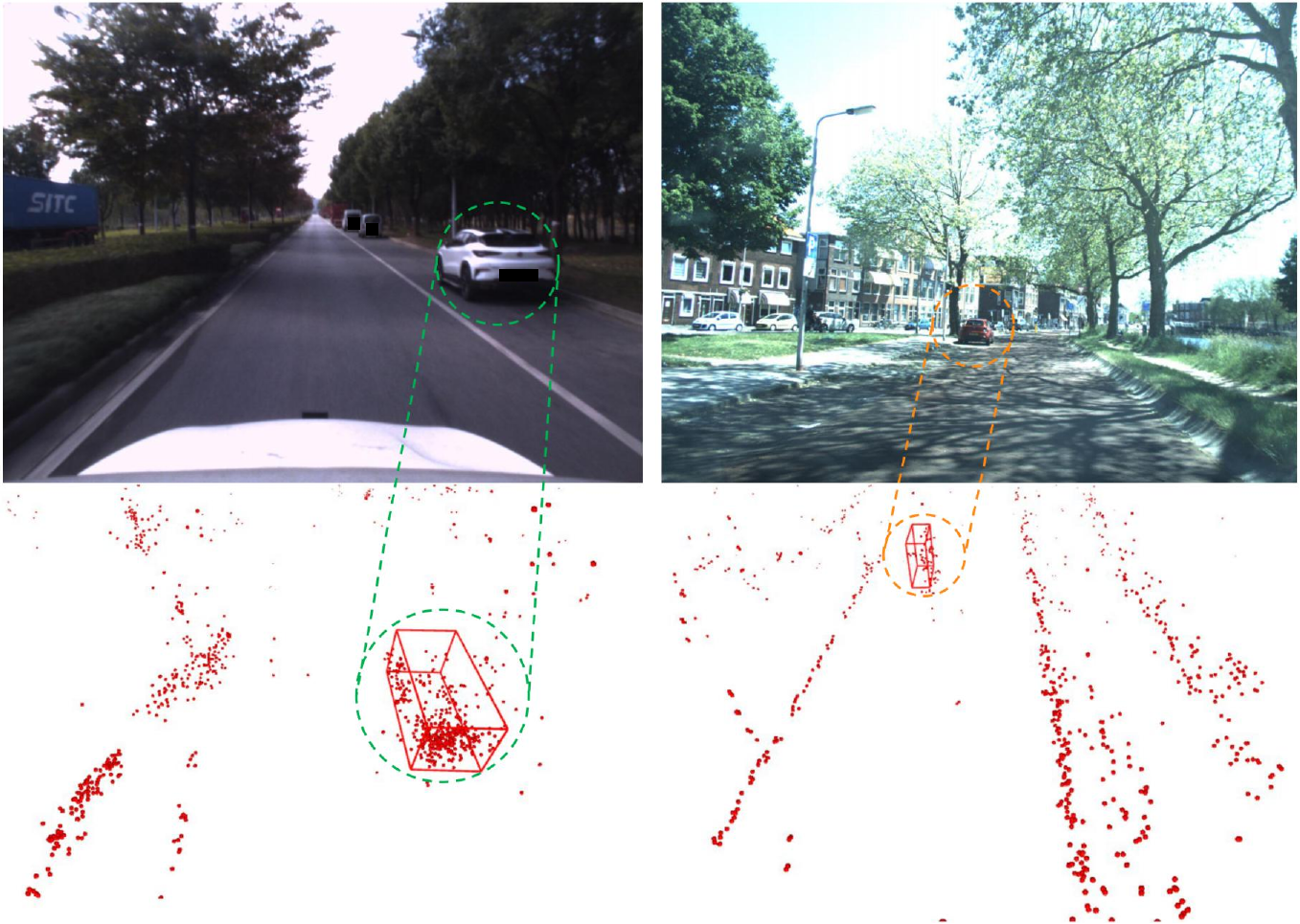}
\caption{
Detection points from the same object tend to be more concentrated. The sub-figure on the left is from 4D radar point cloud provided by TJ4DRadSet \cite{TJ4DRadSet_dataset} using Oculii Eagle under enhanced mode; The sub-figure on the right is from 4D radar point cloud provided by View-of-Delft (VoD) dataset \cite{VoD_dataset} using ZF FRGen21 with 5 scans accumulation.}
\label{intro}
\end{figure}

Consequently, there is a pressing need to develop effective and efficient 3D object detection models based on 4D radar data. Remarkable efforts have recently emerged to tackle this issue. 
For example, a self-attention mechanism proposed by \cite{4d_imaging_radar_3d_object_detection_2} extracts global information from the pillarized radar point cloud to enhance the detection accuracy of the network.
RadarMFNet \cite{4d_imaging_radar_3d_object_detection_3} adopts a modified PointPillars \cite{PointPillar} variant to effectively capture spatiotemporal features by integrating information from the current scan with consecutive scans. However, they don't sufficiently tackle the adverse effects of noise points inherent in the 4D radar point cloud data.

To address the challenges of noise mitigation and efficient feature extraction from sparse 4D radar point cloud, we propose a multi-representation feature encoder. Specifically, our approach involves representing the point cloud as a volumetric grid of fixed-sized pillars and extracting local features within each pillar. Furthermore, to overcome the limitations posed by sparse and noisy data, we introduce a novel approach based on kernel density estimation (KDE) \cite{kde} for extracting density features from the 4D radar point cloud. Since the points generated from different objects show different densities, for example, noise points belonging to clutters are randomly and irregularly distributed with lower density, while points from the same object are more concentrated according to a certain pattern as depicted in Fig. \ref{intro}, KDE helps to achieve additional feature which effectively represents the 4D radar point cloud.
By integrating the pillarization method with the KDE method, we propose a multi-representation feature fusion model for 3D object detection based solely on 4D radar data. The contributions of our work are summarized as follows:
\begin{itemize}
\item 
We propose a novel spatial multi-representation fusion (SMURF) model for 3D object detection. This model is built upon single-modal 4D millimeter-wave radar data and utilizes pillarization and KDE techniques to extract multiple feature representations from the radar point cloud. This study represents the first attempt to employ KDE to extract additional statistical features from a 4D millimeter-wave radar point cloud, for 3D object detection. By capturing the density features of a multi-dimensional Gaussian mixture distribution, KDE effectively mitigates the adverse impact of inherent noise and sparsity in the point cloud.
SMURF can be a formidable baseline for future research and applications.
\item 
To validate the effectiveness and generalization ability of SMURF, we conduct experiments on the VoD dataset \cite{VoD_dataset} and the TJ4DRadSet dataset \cite{TJ4DRadSet_dataset}. The results demonstrate that SMURF surpasses the performance of the latest 4D imaging radar-based approaches and achieves comparable performance to the state-of-the-art approach using 4D radar and camera fusion.
\item Comparable to the inference speed of real-time detection works \cite{real-time 1}\cite{real-time 2}\cite{real-time 3} which typically achieve inference speeds ranging from 10 to 30 FPS, the inference speed of the SMURF model implemented in the MMDetection3D \cite{mmdetection3d} framework is no less than 23 frames per second (FPS). Therefore, SMURF exhibits promising applicability in practical engineering scenarios.
\end{itemize}

The subsequent sections of this paper are organized as follows: Section \Rmnum{2} presents a thorough review of object detection methodologies employing conventional automotive radar and 4D radar. Section \Rmnum{3} provides a detailed description of the proposed SMURF model. In Section \Rmnum{4}, the performance of the SMURF model is evaluated on the VoD and TJ4DRadSet datasets, demonstrating its effectiveness and efficiency in object detection using 4D radar data. Finally, Section \Rmnum{5} concludes the paper by summarizing the key findings and suggesting potential directions for future research.

%==============================================================
\section{Related Works}\label{relatedwork}

In this section, we will begin by reviewing  object detection methods based on conventional automotive radar. Subsequently, we will delve into the realm of 3D object detection utilizing 4D millimeter wave imaging radar.

\subsection{Conventional Automotive Radar-based Object Detection}
Typically, common public radar datasets provide data in the form of point cloud. The sparsity in traditional radar point cloud poses a significant challenge for extracting sufficient information using grid-based methods, while methods directly applied to point cloud can be beneficial in achieving a wider receptive field and valuable features.
For instance, PointNet \cite{pointnet} is employed in \cite{ordinary_radar_based_3d_bev_object_detection_point_processing_1} to extract point-wise features, which are then used for segmentation. 
Similarly, clustering with semantic information is utilized in \cite{radar_instance_segmentation_1} for radar point cloud instance segmentation, and the same approach is combined with contrastive learning in ~\cite{radar_instance_segmentation_2} to solve the lack of radar point annotations. Moreover, a PointNet-based network is employed in \cite{ordinary_radar_based_3d_bev_object_detection_point_processing_2} for object classification, segmentation, and 2D bounding box prediction on the bird's-eye view (BEV), whereas 3D bounding box prediction is addressed in \cite{ordinary_radar_based_3d_bev_object_detection_point_processing_3}.
Furthermore, some studies try to exploit the semantic information in the inter-point relationships.
For instance, studies such as \cite{ordinary_radar_based_3d_bev_object_detection_graph_1}\cite{ordinary_radar_based_3d_bev_object_detection_graph_2} adopt graph convolutional networks (GCN) by grouping input points and adding density features to each point group based on a predefined Euclidean distance threshold. Similarly, GCN is also utilized in \cite{ordinary_radar_based_3d_bev_object_detection_graph_3} before grid-based processing, and the efficacy of clustering the point cloud before feature extraction is demonstrated in \cite{ordinary_radar_based_3d_bev_object_detection_clustering_classification_1}. Besides, \cite{RadarGNN} also proposes a GCN method for achieving transformation invariance, which is robust to unseen scenarios.

However, the information contained within the point cloud is often diminished due to the involvement of complex digital signal processing (DSP) in their production. In contrast, utilizing range-azimuth-Doppler (RAD) tensors collected by radar can provide more comprehensive information for object detection tasks. For example, RADDet \cite{ordinary_radar_based_3d_bev_object_detection_heatmap_4} directly adopts RAD tensors as the model input, while Major \textit{et al.}~\cite{ordinary_radar_based_3d_bev_object_detection_heatmap_1} and RAMP-CNN~\cite{ordinary_radar_based_3d_bev_object_detection_heatmap_2} extract features from the range-azimuth (RA), RD, and azimuth-Doppler (AD) maps. In addition, temporal information  is incorporated to improve detection capabilities in \cite{ordinary_radar_based_3d_bev_object_detection_heatmap_7} and \cite{ordinary_radar_based_3d_bev_object_detection_heatmap_8}. Furthermore, 
the CNN-based model proposed by Cozma \textit{et al.}~\cite{ordinary_radar_based_3d_bev_object_detection_heatmap_12} utilizes a network structure search algorithm for target classification.

Although the approaches mentioned above demonstrate outstanding performance by utilizing the complete information from RAD tensors, memory utilization remains a significant concern, necessitating the development of lightweight alternatives. In this regard, some studies focus on subsets of RDA data.
For instance, RTCnet \cite{ordinary_radar_based_3d_bev_object_detection_heatmap_3} utilizes a small subset from the entire RAD tensor, fed into a CNN to estimate the semantic information. Furthermore, the target recheck system 
\cite{ordinary_radar_based_3d_bev_object_detection_heatmap_9} based on YOLO \cite{yolo}, T-RODNet \cite{ordinary_radar_based_3d_bev_object_detection_heatmap_10}, Patel \textit{et al.}~\cite{ordinary_radar_based_3d_bev_object_detection_heatmap_11}, and DAROD~\cite{ordinary_radar_based_3d_bev_object_detection_heatmap_6} take only two dimensions of RAD data, and the uncertainty estimation is incorporated into bounding box estimation in \cite{ordinary_radar_based_3d_bev_object_detection_heatmap_5}. Furthermore, the RA map can be fused with data obtained from other sensor modalities. For example, RaLiBEV \cite{Radar_LiDAR_fusion_object_detection_RaLiBEV} combines the radar RA heatmap with LiDAR point cloud data to estimate bounding boxes in the BEV.

Despite the existence of conventional radar-based object detection algorithms mentioned in this section, they suffer from a lack of height measurement. This restricts their capacity to provide comprehensive information and hampers their ability to predict 3D bounding boxes. Consequently, the emergence of 4D millimeter-wave radar, which offers height measurement functionality, has captured  %significant 
the attention from both the academic and industrial communities.

\subsection{4D Imaging Radar based 3D Object Detection}
Similar to conventional automotive radar, direct learning methods applied to the entire point cloud can also be extended to 4D radar data. Radar
Transforme~\cite{4D_imaging_radar_classification} is proposed for radar point cloud classification based on the Transformer model. The network takes the multi-dimensional information of the radar point cloud as input, passes it through an encoding layer dominated by multi-layer attention modules, and performs classification based on the learned features.

However, with the advancement of 4D millimeter-wave radar technology, 4D radar point cloud are progressively becoming denser than conventional radar point cloud. Consequently, applying methods that directly learn point features from denser data types, such as the entire point cloud, increases computational complexity. Researchers have started exploring grid-based learning methods as an alternative to tackle this challenge. For instance, the point cloud is divided into pillars and global features are then extracted from the pillars using the attention mechanism in \cite{4d_imaging_radar_3d_object_detection_2}. Similarly, pillar-based detection is also employed in \cite{4d_imaging_radar_3d_object_detection_3}. Prior to pillarization, they first estimate the velocity of the ego vehicle to compensate for the relative radial velocity information in the point cloud and obtain absolute radial velocity.

Furthermore, the fusion of 4D imaging radar with other sensor modalities such as cameras, has enhanced 3D object detection performance. For instance, both the front view (FV) and BEV images are generated from radar point cloud data in \cite{4d_imaging_radar_3d_object_detection_1}, which are then combined with RGB images. Two fusion modules are used in \cite{camera_radar_fusion_3d_object_detection_5} to extract and fuse features from  image and point cloud, and the orthographic feature transform (OFT) \cite{OFT} is used in \cite{4d_imaging_radar_camera_3d_object_detection_rcfusion} for sampling the image pixels. LXL \cite{4d_imaging_radar_camera_3d_object_detection_LXL} utilizes predicted depth distribution maps and radar 3D occupancy grids as auxiliary elements to transform the multi-level perspective view maps into a BEV map. In addition to camera images, LiDAR point cloud is also commonly used for multi-modal fusion with 4D imaging radar. For example, InterFusion \cite{4d_imaging_radar_lidar_3d_object_detection_1} fuses the features of radar and LiDAR point cloud using a self-attention-based method, allowing the network to focus on relevant features while ignoring irrelevant ones. Besides, $\mathrm{M^{2}}$-Fusion \cite{4d_imaging_radar_lidar_3d_object_detection_2} first pillarlizes the point cloud from LiDAR and 4D radar, then utilizes a self-attention mechanism to learn features from both modalities and exchange the intermediate layer information.

While there have been several methods, they do not adequately tackle the issue of noise present in the 4D radar point cloud. Additionally, multi-modal fusion methods typically require strict synchronization and a lot of computational power, which are problematic. Therefore, in this work, we seek to address these issues and develop a novel approach that uses fewer computational resources and aims to reduce the impact of noise while extracting distinctive features from the single sparse 4D radar point cloud only.

\subsection{Other Applications with 4D Imaging Radar}

In addition to object detection, 4D millimeter-wave radar finds extensive utility across various domains. For instance, Li \textit{et al.}~\cite{other works on 4d radar 1} propose a SLAM framework called 4DRaSLAM, which addresses the challenges of simultaneous localization and mapping. It filters out ghosts and random points from the 4D radar point cloud, utilizes static points and their Doppler velocity information to estimate self-velocity, and achieves improved pose estimation. While 4DRadarSLAM~\cite{other works on 4d radar 2}, comprising three modules, enhances SLAM performance by incorporating the probability distribution of each point and other factors.
Besides SLAM, 4D radar can be employed for other tasks as well. For instance,  Ding \textit{et al.}~\cite{other works on 4d radar 3} employ self-supervised learning to estimate scene flow from 4D radar point cloud, enhancing support for downstream motion segmentation tasks. While  CMFlow \cite{other works on 4d radar 4} leverages cross-modal supervision to perform scene flow estimation based on 4D radar. As a result, it is evident that 4D radar possesses extensive application prospects and potential.

\textcolor{blue}{}

\section{The Proposed Method}   
\label{proposed methods}

\begin{figure*}[!htb]
\centering
\includegraphics[width=\textwidth]{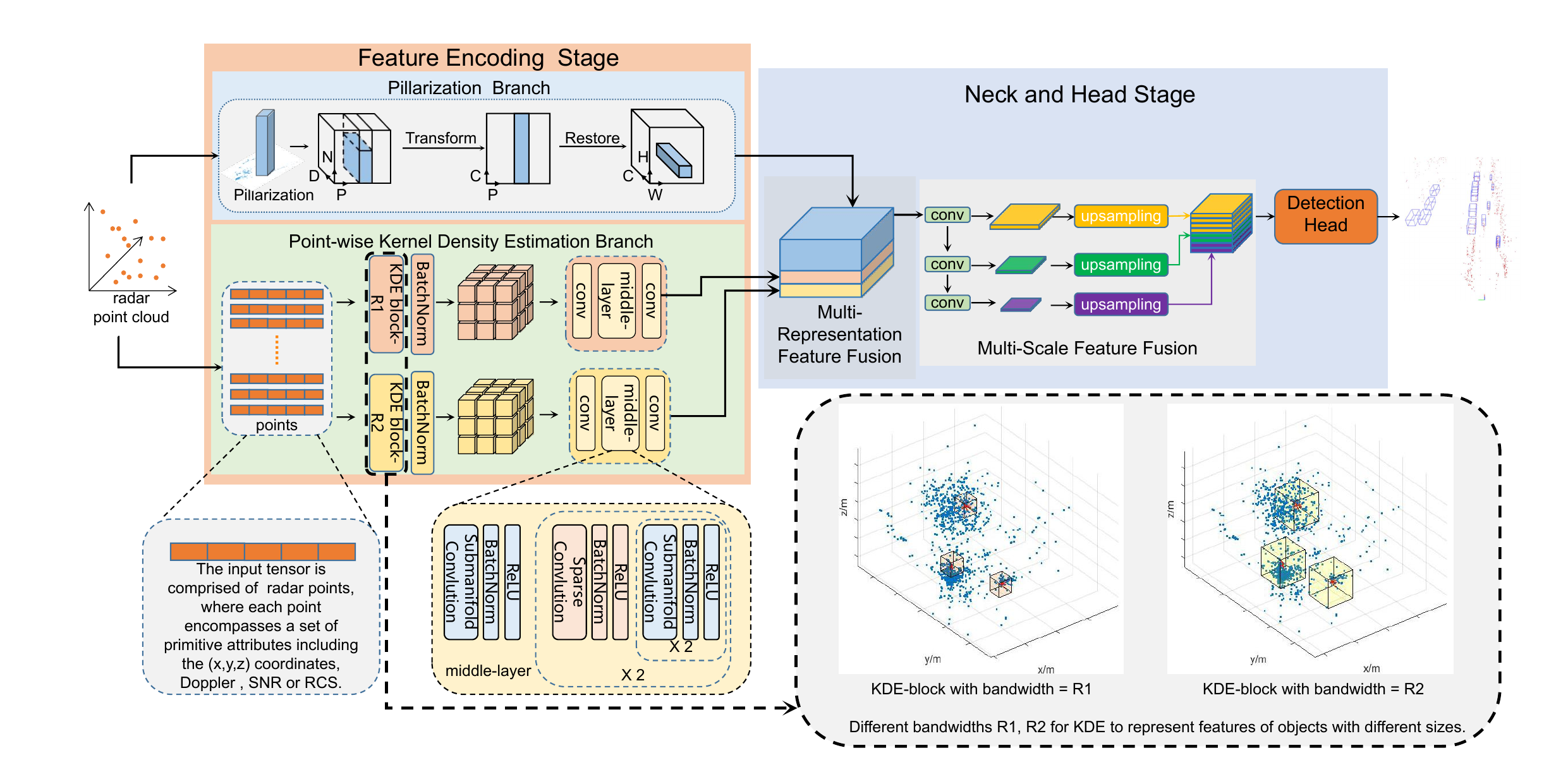}
\caption{The pipeline of our proposed SMURF 3D object detector consists of two stages, taking point cloud as the input data. The feature encoding stage consists of two branches: The first branch is the pillarization branch, responsible for extracting features by partitioning the point cloud space into pillars, while the second branch conducts KDE on the tensor of the point cloud with different bandwidths, normalizes it, and then partitions the resulting point cloud into voxels to extract features. The combination of these two branches enables the extraction of multi-representation features of the point cloud. In the following neck and head stage, the multi-representation features are fused, followed by further feature extraction and encoding using a multi-scale network approach. Finally, an anchor-box-based detection head is employed to generate prediction results for 3D bounding boxes, object categories, confidence scores, and other relevant attributes.}
\label{CP_Process}
\end{figure*}

In this section, we provide detailed descriptions of SMURF for object detection based on the fusion of multiple feature representations extracted from a single modal 4D radar. The structure of SMURF is illustrated in Fig.~\ref{CP_Process}. The point cloud obtained from the 4D millimeter-wave radar is characterized by sparsity and a higher noise level than sensors such as LiDAR. These characteristics pose challenges to effective feature extraction using a single-representation approach. To address this, we propose a novel method for multi-representation feature encoding with two branches.

The proposed SMURF model consists of two stages. The first stage is the feature encoding stage. It incorporates the pillarization branch and the point-wise KDE branch. They are responsible for extracting point cloud pillarization features and KDE features, respectively. The second stage is the neck and head stage. The neck includes the multi-representation feature fusion (MRFF) module and the multi-scale feature fusion (MSFF) module, where MRFF fuses multi-representation features, and MSFF is responsible for further extraction and integration of multi-scale features. 
Finally, an anchor-box-based detection head is employed to generate prediction results, including 3D bounding boxes and object categories.

%-------------------------------------------
\subsection{Feature Encoding Stage: Pillarization Branch}
In this branch, we follow the methodology proposed in PointPillars \cite{PointPillar}. First, we partition the 4D radar point cloud into pillars along $X$ and $Y$ axes. Let $P$ denote the number of non-empty pillars. 
The number of points within each non-empty pillar may vary and can be assigned a reference value $N$. For pillars containing more than $N$ points, we extract a random subset of $N$ points. For pillars with fewer than $N$ points, we artificially add additional points with zero values to ensure the total number of points equals $N$. 
Next, we extend the features of each point $p$ to dimension $D$. The extended feature representation can be defined as follows:
\begin{equation}
    D=[D_{raw},x_c,y_c,z_c,x_p,y_p,z_p],
\end{equation}
where $D_{raw}$ represents the original features of the point, and $[x_c,y_c,z_c]$ represents the distance of each point with respect to the centroid of pillars. $[x_p,y_p,z_p]$ represents the distance of each point with respect to the geometric center of pillars, in contrast to PointPillars \cite{PointPillar}, which only considers $[x_p,y_p]$. As a result, we obtain a sparse $(D,P,N)$ tensor that encapsulates the point cloud information in a structured manner. 

The $(D, P, N)$ sparse tensor, representing the point cloud, undergoes a mapping process to transform it into a $(C, P, N)$ space along the $D$ axis, where $C$ represents the mapped features. This transformation is achieved by applying linear, batch normalization, and ReLU layers. Subsequently, a max pooling operation is performed along the $N$ dimension, resulting in a $(C,P)$ tensor. This tensor is then passed through a multi-layer perceptron (MLP) layer, yielding a tensor with a shape of $(C_{1},P)$. Finally, to restore the tensor to a $(C_{1},H,W)$ feature map, the position encoding of the $(C_{1},P)$ tensor is combined with the size of the dense tensor.

%--------------------------------------------
\subsection{Feature Encoding Stage: Point-wise Kernel Density Estimation Branch}

\begin{figure}[htbp]
\centering
\includegraphics[width=\linewidth,scale=1.00]{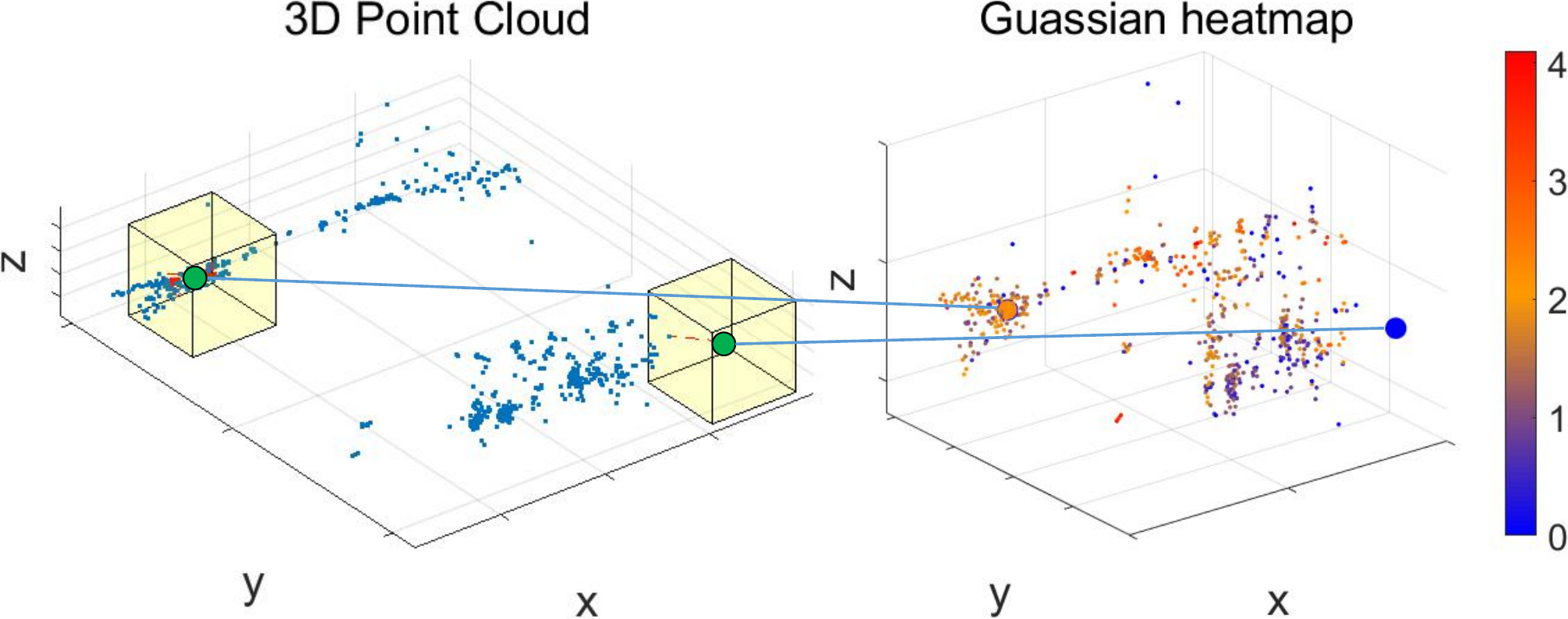}
\caption{The KDE block operates by extracting density features from point cloud data to obtain a multi-dimensional Gaussian mixture distribution heatmap.
The sub-figure on the left demonstrates the process of density feature extraction using a local perspective of point cloud data (shown in blue). Each point computes its kernel function based on the points within a certain distance range (shown as an yellow cube) centered around that point. The sub-figure on the right visualizes the local resulting Gaussian mixture distribution heatmap. Heat values corresponding to two points (shown in green) in the left sub-figure increase with point density.}
\label{kde_block}
\end{figure}

To enhance the semantic features of the point cloud, we introduce the non-parametric KDE method in the feature encoding stage. In this point-wise kernel density estimation branch, we consider an input tensor of shape $(N_{p},C_{raw})$, where $N_{p}$ represents the number of radar points, and $C_{raw}$ denotes the $(x,y,z)$ coordinates as well as other raw features associated with each radar point. To extract density features, we employ the KDE block as illustrated in Fig.~\ref{kde_block}, which is composed of a multi-dimensional Gaussian function. For each point, the KDE block calculates the differences on multi-feature dimensions between itself and the surrounding points within a specific range. These differences are then mapped using a Gaussian kernel function. The resulting values are multiplied across the multi-feature dimensions and subsequently summed and averaged to obtain an estimation of the density for that particular point. Specifically, the density feature $\rho(p)$ of each point $p$ is defined by taking into account the $(x, y, z)$ coordinates and Doppler information (as an example):
\begin{align}
    \rho(p)=\frac{1}{M_{p}R^3}\sum_{i=1}^{M_{p}}\prod_{t\in(x,y,z,Dop)}K_{R}(t,t_{i}),
\end{align}
\begin{align}
    \text{subject to} \ \ &\left\{
    \begin{aligned}
    |x-x_i|\leq R\\
    |y-y_i|\leq R\\
    |z-z_i|\leq R ,\\
    \end{aligned}
    \right.
\end{align}
where $Dop$ represents the Doppler feature of points, and $M_{p}$ represents the number of other points $p_i$ with the $(x_{i},y_{i},z_{i})$ coordinates within a certain distance from $p$. The kernel function $K_{R}(\cdot,\cdot)$ is defined as Gaussian kernel:
\begin{equation}
\begin{aligned}
    K_{R}(t,t_{i})=e^{{||\frac{t-t_{i}}{R}||}^2}, && t\in(x,y,z,Dop,\ldots).
\end{aligned}
\end{equation}
Since Gaussian kernel function is used, the KDE of radar detection points reflects the multi-dimensional Gaussian mixture distribution of the points, as shown in Fig.~\ref{kde_block}.

Furthermore, the bandwidth for regulating the influence range of the kernel function, $R$, is important in controlling the degree of smoothing versus oscillation of the estimated density function. In this branch, we have incorporated different bandwidth sizes for KDE to extract features from objects of varying sizes in the 4D radar point cloud. 
The choice of bandwidth size influences the shape of the kernel function used in KDE, which, in turn, affects the sensitivity to objects of different sizes. Specifically, a smaller bandwidth corresponds to a kernel function with a narrower window, resulting in higher sensitivity to smaller objects within the point cloud. Conversely, a larger bandwidth leads to a flatter kernel function, allowing for a wider window and better capturing of features related to larger objects in the point cloud. 

After KDE, the shape of the point cloud tensor is transformed to $(N_{p},C_{2})$, where $C_{2}=1$. It is important to note that the 4D radar point cloud often contains significant noise. When using the input tensor composed of all points for KDE calculation, the presence of noise can adversely affect the density estimation results. Consequently, density values may become non-zero for all points, including those noise points from clutters. To address this issue, it is desirable to assign negative density values to isolated noise points. This normalization step helps ensure that the density characteristics of the input tensor, which represents the 4D radar point cloud, are appropriately adjusted. For each point $p$ with density feature $\rho$, the normalization is performed by
\begin{equation}
    \mu=\sum_{i=1}^{N_{p}}\rho_{i},
\end{equation}
\begin{equation}
    \sigma^{2}=\frac{1}{N_{p}}\sum_{i=1}^{N_{p}}(\rho_{i}-\mu)^{2},
\end{equation}
\begin{equation}
    \rho_{bi}=\frac{\rho_{i}-\mu}{\sqrt{\sigma^{2}+\epsilon}},
\end{equation}
where $\mu$ and $\sigma$ represent the mean value and variance of the density feature of the point, respectively; and $\epsilon$ is a parameter set to prevent division error caused by zero variance. Based on the normalization of every point $p$, its corresponding density feature $\rho_{b}$ can be obtained. 
 
\begin{figure}[tbp]
\centering
\subfigure[]{\includegraphics[width=0.45\linewidth]{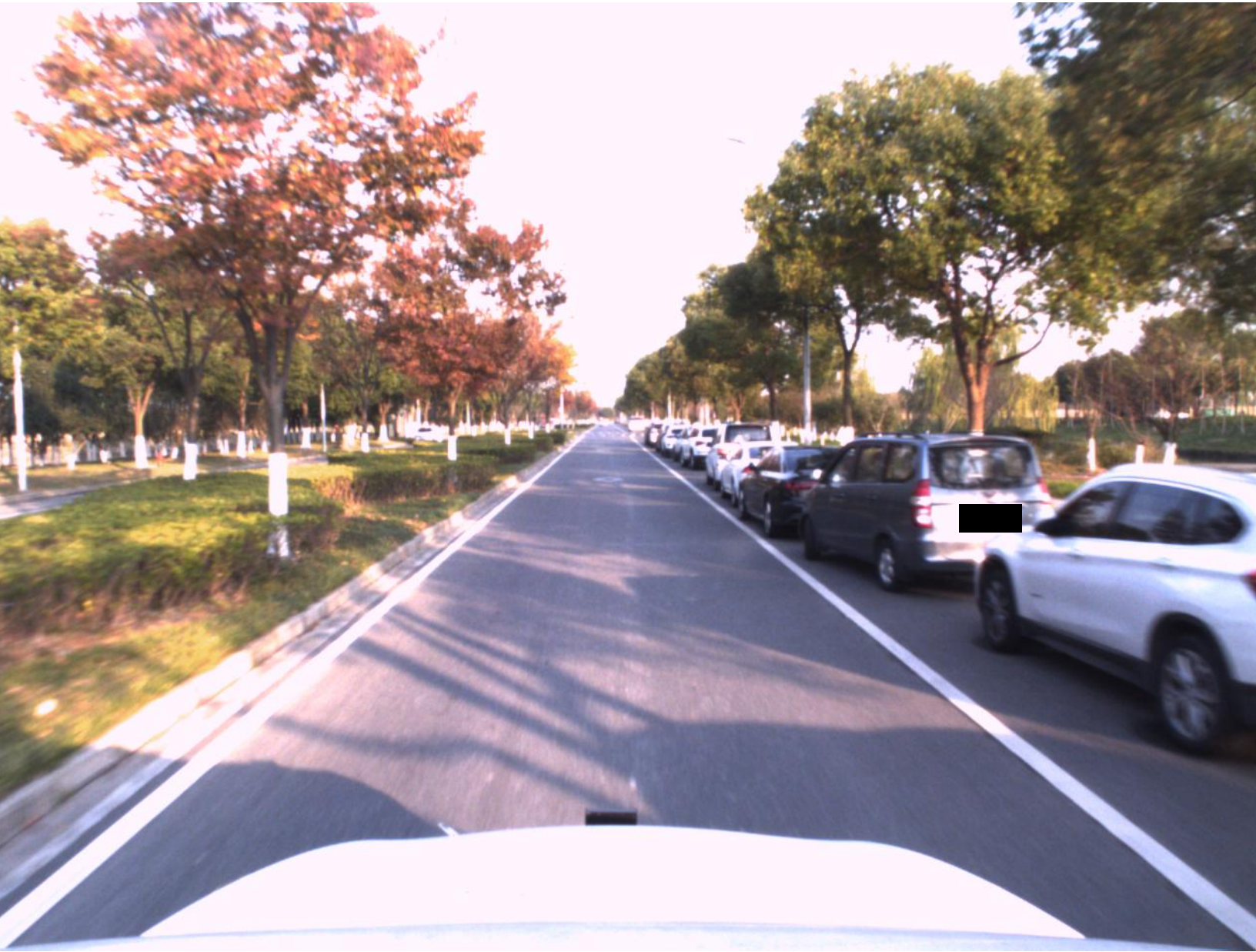}}
\subfigure[]{\includegraphics[width=0.5\linewidth]{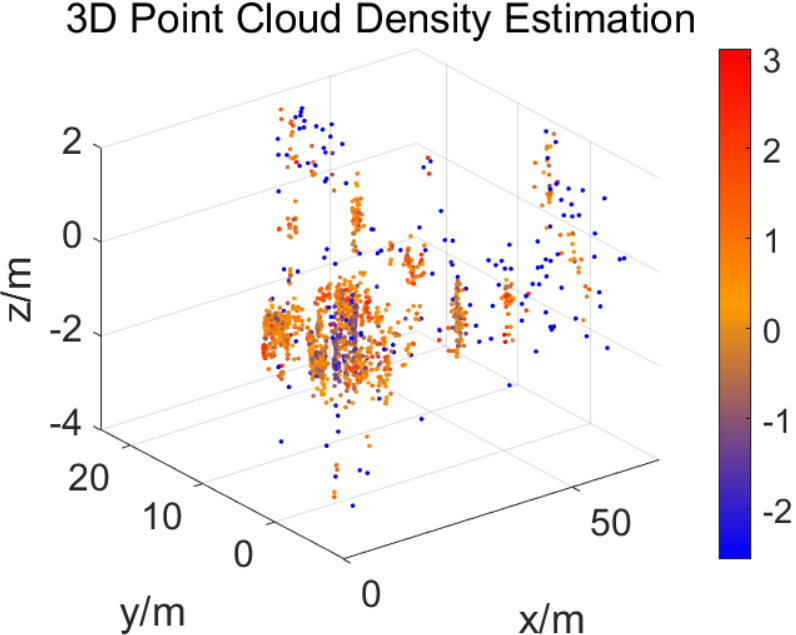}}\\
\subfigure[]{\includegraphics[width=0.435\linewidth]{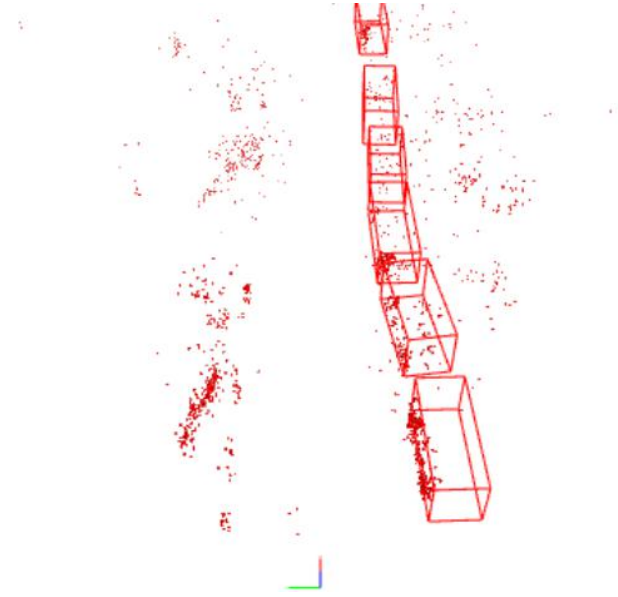}}
\subfigure[]
{\includegraphics[width=0.5\linewidth]{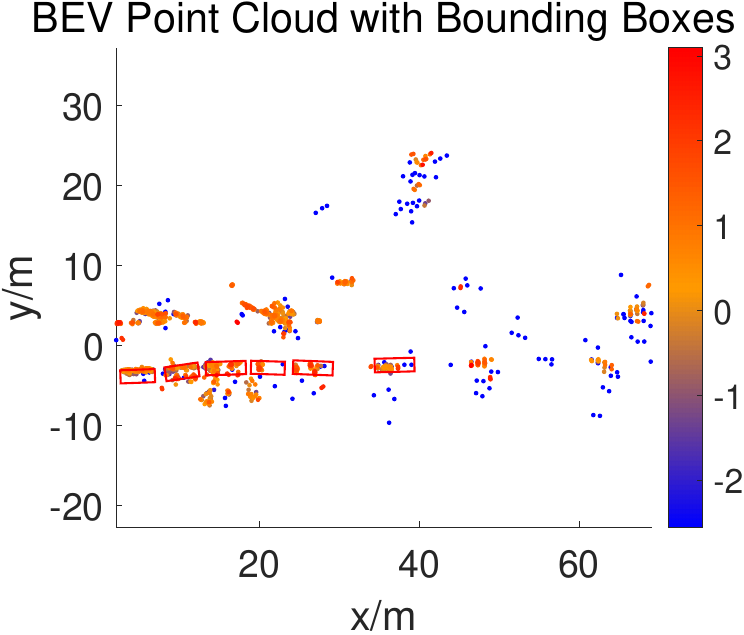}}
\caption{The denoising effect of the KDE method. Here we use a scan of radar and image data from the TJ4DRadSet dataset as an example. (a) shows the image captured from the synchronized camera, while (b) is the 3D point cloud heatmap after KDE and normalization. (c) displays the ground truth bounding boxes within the scene, highlighted in red, with the red points indicating the original point cloud. (d) displays the ground truth bounding boxes along with a heatmap of the point cloud on the BEV plane. It is evident that isolated noise  points, which do not fall within the ground truth bounding boxes, exhibit lower density values. By assigning lower density characteristics to these points, KDE helps mitigate the effect of noise points on the network.}
\label{kde_heatmap}
\end{figure}

Noise points in the 4D radar point cloud often exhibit a random distribution and isolation phenomenon, resulting in lower density values than genuine signal points. The network's sensitivity to noise points is reduced by performing feature extraction through KDE, leading to improved detection performance, as illustrated in Fig.~\ref{kde_heatmap}.  Moreover, the 4D radar point cloud is sparse, while points belonging to the same object tend to be relatively concentrated. By employing KDE, valuable information regarding the spatial distribution and concentration of points can be effectively extracted, enhancing the discriminative power of the network.

To further extract 3D spatial features from point cloud, the voxelization method \cite{second} is introduced in this branch. Following the normalization process, the point cloud is divided into voxels, resulting in a tensor of shape $(C_{2}, D, H, W)$, where $D$, $H$ and $W$ are the depth, height, and width of the 3D pseudo-image, respectively. Subsequently, the middle encoding layers are employed to extract more discriminative features.
These layers primarily consist of convolutional layers and a middle-layer (ML) block. The ML block is constructed using a heterogeneous combination of 3D submanifold sparse convolutional \cite{submconv} and sparse convolutional \cite{sparseconv} layers, normalization layers, and ReLU layers. The combination of these layers facilitates the extraction of higher-level features from the voxelized point cloud data. Notably, 3D sparse convolution is particularly advantageous in reducing redundant computations and inference time. 
By selectively computing features only for active voxels, it effectively exploits the sparsity of the data while retaining significant features. 

Finally, we obtain pseudo-images with shapes $(C_{2,1},H, W)$ and $(C_{2,2},H,W)$ from KDE branch, where $C_{2,1}$ and $C_{2,2}$ are equal, representing the density features of objects with different sizes.
%--------------------------------------------
\subsection{Neck and Detection Head}

The neck network of our model comprises two key components: the MRFF module and the MSFF module.

In the MRFF module, the pseudo-images obtained from the pillarization branch with shape $(C_{1},H,W)$ are combined with the pseudo-images from the KDE branch, which have shapes $(C_{2,1},H, W)$ and $(C_{2,2},H,W)$. To prevent overfitting of features and simplify the model structure, we use the fusion method that directly concatenates the features along the channel dimension. This fusion process results in a pseudo-image with the shape $(C_{f},H,W)$, where $C_{f}=C_{1}+C_{2,1}+C_{2,2}$.

The MSFF module we adopted is from \cite{fpn}. To effectively detect detection targets with multiple categories such as cars and pedestrians, the MSFF module applies different numbers of convolutional layers to obtain multi-scale feature maps. This process increases the network's receptive field, allowing it to capture both local and global features. The obtained multi-scale feature maps have different shapes, namely $(C_{m1},H,W)$, $(C_{m2},H/2,W/2)$ and $(C_{m3},H/4,W/4)$, respectively. These maps represent features at different scales, with the spatial dimensions being halved at each scale. By employing convolutional layers at different scales, the network can capture information at various levels of detail. To ensure compatibility for subsequent operations, the multi-scale feature maps are upsampled to the same size. Following the upsampling process, the feature maps are concatenated along the channel dimension, resulting in a feature map with the shape  $(C_{m},H,W)$, where $C_{m}=C_{m1}+C_{m2}+C_{m3}$.

Finally, to generate predictions for object detection, our proposed SMURF model incorporates an anchor-box-based single-stage detection head, similar to the approach used in PointPillars \cite{PointPillar}. The network refines the predicted bounding boxes by computing the loss function. This loss function evaluates the disparity between the predicted bounding boxes and the ground truth bounding boxes.

The loss function includes the 3D bounding box regression loss $L_{bbox}$ inspired by \cite{bbox loss}, the classification loss $L_{cls}$ from \cite{cls loss}, and the direction loss $L_{dir}$ used in PointPillars \cite{PointPillar}. The 3D bounding box regression loss $L_{bbox}$ represents the difference between the ground truth bounding box $bbox$ and the predicted bounding box $bbox^{p}$, where $bbox=[x^{g},y^{g},z^{g},w^{g},l^{g},h^{g},\theta^{g}]$ and $bbox^{p}=[x^{p},y^{p},z^{p},w^{p},l^{p},h^{p},\theta^{p}]$. Thus $L_{bbox}$ is defined by
\begin{equation}
    L_{bbox}=\frac{1}{N_{pos}}\sum_{i=1}^{N_{pos}}\sum_{t_{i}\in[x_{i},y_{i},z_{i},w_{i},l_{i},h_{i},\theta_{i}]}SmoothL1(\delta t_{i}),
\end{equation}
\begin{equation}
    SmoothL1(x)=\left\{
    \begin{aligned}
    & \text{} && 0.5x^{2}, & \text{} && \mathrm{if}\ \  |x|<1\\
    & \text{} && |x|-0.5,  & \text{} && \mathrm{otherwise},\\
    \end{aligned}
    \right.
\end{equation}
\begin{equation}
    \delta t =\left\{
    \begin{aligned}
    & \text{} && \frac{t^{g}-t^{p}}{\sqrt{(w^{p})^{2}+(l^{p})^{2}}}, & \text{} && \mathrm{if}\ \  t\in[x,y]\\
    & \text{} && \frac{t^{g}-t^{p}}{h^{p}}, & \text{} && \mathrm{if}\ \  t=z\\
    & \text{} && \log(\frac{t^{g}}{t^{p}}), & \text{} && \mathrm{if}\ \  t\in[w,l,h]\\
    & \text{} && \sin({t^{g}-t^{p}}), & \text{} && \mathrm{if}\ \  t=\theta,\\
    \end{aligned}
    \right.
\end{equation}
where $N_{pos}$ represents the number of positive samples, and positive and negative samples are divided based on Intersection over Union (IoU) values within ground truth bounding boxes.

The classification loss $L_{cls}$ is defined by
\begin{equation}
    L_{cls}=-\frac{1}{N_{pos}}\sum_{i=1}^{N_{pos}}\sum_{j=1}^{N_{cls}}\sigma_{i,j}\alpha_{j}(1-p_{i,j})^{\gamma}\log{p_{i,j}},
\end{equation}
\begin{equation}
    \sigma_{i,j}=\left\{
    \begin{aligned}
    & 1,\ \ & \mathrm{if}\ \ C(S_{i})=j\\
    & 0, & \mathrm{otherwise},\\
    \end{aligned}
    \right.
\end{equation}
where $S_{i}$ represents the current sample, and $C(\cdot)$ is used to determine the current object category. $N_{cls}$ represents the number of object classes; $\alpha_{i,j}$ is the weight factor; and $p$ is the probability of an anchor belonging to a certain category.

Finally, the total loss $L$ is defined by
\begin{equation}
    L=\beta_{1}L_{bbox}+\beta_{2}L_{cls}+\beta_{3}L_{dir},
\end{equation} 
where $L_{dir}$ represents the loss between the predicted target direction and the ground truth direction; and $\beta_{1}, \beta_{2}$, and $\beta_{3}$ represent the weight of each loss value.

%===============================================
\section{Experiments and Analysis}
\label{result}

%-------------------------------------------
\subsection{Dataset and Evaluation Metrics}
\label{metrics}

Point cloud data is a commonly provided format in publicly available 4D radar datasets. For instance, the Astyx 6455 HiRes millimeter-wave radar is used to collect data in \cite{Astyx_HiRes_2019_dataset}, providing  ground truth annotations for seven classes, including cars and buses, but with only 500 scans. In contrast, the VoD dataset \cite{VoD_dataset} provides synchronized data of images, LiDAR point cloud, and 4D radar point cloud, comprising 8,600 scans and 3D bounding box annotations for over 26,000 pedestrians, 10,000 cyclists, and 26,000 cars.
The VoD dataset includes three types of 4D radar point cloud: single-scan, three-scan, and five-scan. However, one limitation is the lack of common driving scenarios, such as highways. As a valuable complement, TJ4DRadSet \cite{TJ4DRadSet_dataset} offers diverse driving scenarios with different lighting conditions and traffic types. It provides well-annotated 3D bounding boxes and track IDs for more than 16,000 cars, 5,300 trucks, 4,200 pedestrians, and 7,300 cyclists, encompassing 7,757 scans of data. 

On another note, raw radar tensors preserve more comprehensive information compared to radar point cloud data. RADIal \cite{RADIal_dataset} provides ADC data along with the RAD tensor and RA map. It consists of approximately 25,000 scans and offers 2D bounding box labels for 9,550 vehicles. However, the absence of 3D bounding box labels may limit its applicability for certain tasks. K-Radar dataset \cite{K-Radar_dataset} includes the raw ADC data and the Range-Azimuth-Elevation-Doppler (RAED) tensor. This dataset encompasses 35,000 scans with annotated 3D bounding boxes for 93,300 objects, covering various types of traffic participants, such as sedans, pedestrians, and buses.  Furthermore, it encompasses challenging driving conditions in adverse weather conditions (e.g., fog, rain, and snow) and various road structures, making it suitable for studying 4D radar robustness. However, in practical applications, the size of ADC data or RAED tensor is too large for the transmission capacity of the controller area network-bus (CAN bus) protocol on vehicle, making it impossible to employ them in neural networks for 3D object detection. Besides, commercially available 4D millimeter-wave radars commonly offer only point cloud outputs. Although the K-Radar dataset provides point cloud with only $(x,y,z)$ features, the absence of Doppler information is unacceptable for a 4D millimeter-wave radar point cloud. To ensure algorithm generalization, we exclude this dataset from consideration.

In this study, we evaluate the generalization capability of the proposed SMURF method using the VoD and TJ4DRadSet datasets.
For the VoD dataset, we utilize the original features extracted from the point cloud, which consist of seven dimensions. The feature vector is represented as:
\begin{equation}
    D_{raw}=[x,y,z,RCS,v_r,v_{rc},\tau],
\end{equation}
where $(x,y,z)$ denote the coordinates of the radar points, $RCS$ represents the radar signal reflection-intensity, $v_{r}$ is the radial Doppler velocity relative to the ego vehicle, $v_{rc}$ is the absolute Doppler velocity, and $\tau$ indicates the time ID indicating which scan the point belongs to. On the other hand, for the TJ4DRadSet dataset, we also utilize the original features extracted from the point cloud, which consist of five dimensions. The feature vector is denoted as:
\begin{equation}
    D_{raw}=[x,y,z,v_r,SNR],
\end{equation}
where $SNR$ denotes the signal to noise ratio of the detection.

For the VoD dataset, we utilize 5-scan accumulated radar point cloud data to evaluate the detection of three distinct object categories: cars, pedestrians, and cyclists. The evaluation metrics employed include 3D average precision ($\mathrm{AP_{3D}}$) values for each object category, as well as the mean 3D AP ($\mathrm{mAP_{3D}}$) and mean BEV AP ($\mathrm{mAP_{BEV}}$) values. These metrics are computed separately for the entire annotated area and driving corridor specified in the official guidelines. Following the official settings, the IoU thresholds used for calculating the performance metrics are set to 0.5 for cars, 0.25 for pedestrians, and 0.25 for cyclists.

On the other hand, for the TJ4DRadSet dataset, we evaluate the detection performance on four object categories: cars, pedestrians, cyclists, and trucks. Similarly, following the settings in \cite{4d_imaging_radar_camera_3d_object_detection_rcfusion}, the IoU thresholds for each category are set to 0.5 for cars, 0.25 for pedestrians and cyclists, and 0.5 for trucks. Unlike the VoD dataset, the TJ4DRadSet dataset allows for the specification of evaluation areas based on the distance to the onboard sensors. In our evaluation, we focus on objects within a range of 70 meters from the radar. The evaluation metrics considered for the TJ4DRadSet dataset include BEV AP ($\mathrm{AP_{BEV}}$) and $\mathrm{AP_{3D}}$ values for each category. The $\mathrm{mAP_{BEV}}$ and $\mathrm{mAP_{3D}}$ values are also crucial evaluation metrics to assess the overall performance of the SMURF method on the TJ4DRadSet dataset.

\subsection{Implementation Details}
\label{Implementation}

\subsubsection{Hyper-parameter Setting}
For the VoD dataset, the radar point cloud has a range interval of 0 to 51.2 meters on the x-axis, $-$25.6 to 25.6 meters on the y-axis, $-$3 to 2 meters on the z-axis. Similarly, for the TJ4DRadSet dataset, the radar point cloud has a range interval of 0 to 69.12 meters on the x-axis, $-$39.68 to 39.68 meters on the y-axis, $-$4 to 2 meters on the z-axis. Pillar dimensions are set to 0.16m in length and width and voxel dimensions are set to 0.16m (length) $\times$ 0.16m (width) $\times$ 0.24m (height) for both datasets.

In the evaluation of the SMURF model on VoD and TJ4DRadSet datasets, KDE is employed with different bandwidth parameters. Specifically, for VoD dataset, two bandwidth values are utilized: $1.5$ meters and $2$ meters. For TJ4DRadSet dataset, bandwidth values of $0.6$ meters and $1$ meter are employed. These bandwidth values are chosen to adapt to the specific characteristics and resolution of each dataset. 

Taking the VoD dataset as an example, after the MRFF stage, the resulting feature map has a resolution of $320\times320$. In the subsequent MSFF stage, a set of multi-scale feature maps is generated using down-sampling rates of 2, 4, and 8. These feature maps are then up-sampled to a resolution of $160\times160$ for the purpose of feature map fusion. As a result, all the feature maps input into the detection head have a consistent resolution of $160\times160$. The selection of resolution in our study is based on two considerations. Firstly, we draw insights from the prior works including Second \cite{second}. Secondly, we take into account the inherent sparsity of radar point cloud. Excessively high down-sampling rates would lead to a substantial loss of informative content in the resulting feature maps.

Moreover, in the object detection process, predefined anchor boxes are utilized in the detection head. For the VoD dataset, anchor boxes are defined for the car, pedestrian, and cyclist classes. The dimensions of the anchor boxes for these classes are (3.9m, 1.6m, 1.56m), (0.8m, 0.6m, 1.73m), and (1.76m, 0.6m, 1.73m), respectively. The $z$ coordinates of the bottom center positions for these anchor boxes are predefined as $-$1.78m, $-$0.6m, and $-$0.6m, respectively. Additionally, the orientation angles of the anchor boxes are predefined to be either 0 degrees or 90 degrees.

Similarly, for the TJ4DRadSet dataset, anchor boxes are defined for the car, pedestrian, cyclist, and truck classes. The anchor box dimensions for these classes are set to (1.84m, 4.56m, 1.70m), (0.6m, 0.8m, 1.69m), (0.78m, 1.77m, 1.60m), and (2.66m, 10.76m, 3.47m), respectively. The $z$ coordinates of the bottom center positions for these anchor boxes are predefined as $-$1.363m, $-$1.163m, $-$1.353m, and $-$1.403m, respectively. The orientation angles of the anchor boxes are also predefined as either 0 degrees or 90 degrees. The parameters about the predefined anchor boxes designed for both datasets are obtained directly from their official configurations.

\subsubsection{Training Setting}
The training of SMURF is processed on a single NVIDIA V100 GPU using MMDetection3D toolbox \cite{mmdetection3d}. The network is trained using the step decay learning rate strategy, with an initial learning rate of 1e-3. The AdamW is employed as optimizer, and the batch size is set to 16. 

In addition, data preprocessing plays a crucial role in preparing the input data for training. It includes data normalization, data augmentation, point filtering, and so on. Point filtering is utilized to eliminate points outside the radar point cloud range and the camera's field of view. Data augmentation techniques, including global scaling and flipping around the x-axis are employed to enhance the model's generalization ability.

\subsection{Ablation Study for SMURF}
\label{ablation_study}

\begin{table*}[htbp]
  \caption{Ablation Study of SMURF on both the VoD validation dataset and TJ4DRadSet test dataset. All the approaches are executed with 5-scans radar detection points on VoD dataset and single-scan radar detection points on TJ4DRadSet dataset.}
  \centering
  \begin{tabular}{c|c|c|c|c|c|c|c|c|c|c}
    \hline
    \multirow{2}*{Method} & \multicolumn{5}{c|}{Entire Annotated Area AP} & \multicolumn{5}{c}{In Driving Corridor AP} \\
    \cline{2-11}
    & Car & Ped & Cyc & $\mathrm{mAP_{3D}}$ & $\mathrm{mAP_{BEV}}$ & Car & Ped & Cyc & $\mathrm{mAP_{3D}}$ & $\mathrm{mAP_{BEV}}$ \\
    \hline
    
    SMURF without pillarization & 34.62 & 13.74 & 32.55 & 26.97 & 31.60 & 66.09 & 22.65 & 56.60 & 48.45 & 52.09\\
    \hline
    
    SMURF without KDE & 41.11 & 37.09 & 65.92 & 48.04 & 55.20 & 70.01 & 48.04 & 86.18 & 68.08 & 69.29\\	 
    \hline
    
    SMURF (\textbf{ours}) (x, y, z, Doppler, RCS)
     & \textbf{42.31} & \textbf{39.09} & \textbf{71.50} & \textbf{50.97} & \textbf{56.77} & \textbf{71.74} & \textbf{50.54} & \textbf{86.87} & \textbf{69.72} & \textbf{72.05}  \\
    \hline
    \hline
    \multirow{2}*{Method} & \multicolumn{5}{c|}{$\mathrm{AP_{3D}}$} & \multicolumn{5}{c}{$\mathrm{AP_{BEV}}$} \\
    \cline{2-11}
    & Car & Ped & Cyc & truck & $\mathrm{mAP_{3D}}$ & Car & Ped & Cyc & truck & $\mathrm{mAP_{BEV}}$ \\
    
    \hline
    SMURF without pillarization & 15.19 & 7.41 & 19.97 & 6.52 & 12.27 & 23.21 & 7.95 & 21.72 & 12.82 & 16.43\\
    
    \hline
    SMURF without KDE & 27.50 & 24.35 & 49.09 & 18.94 & 29.97 & 38.17 & 26.26 & 52.71 & 28.92 & 36.51 \\	 
    \hline

    SMURF (\textbf{ours}) (x, y, z, Doppler, SNR) & \textbf{28.47} & \textbf{26.22} & \textbf{54.61} & \textbf{22.64} & \textbf{32.99} & \textbf{43.13} & \textbf{29.19} & \textbf{58.81} & \textbf{32.80} & \textbf{40.98} \\
    \hline
  \end{tabular}
  \label{ablation}
\end{table*}

\begin{figure*}[htbp]
  \centering
  \subfigure[]{\includegraphics[width=0.2\linewidth]{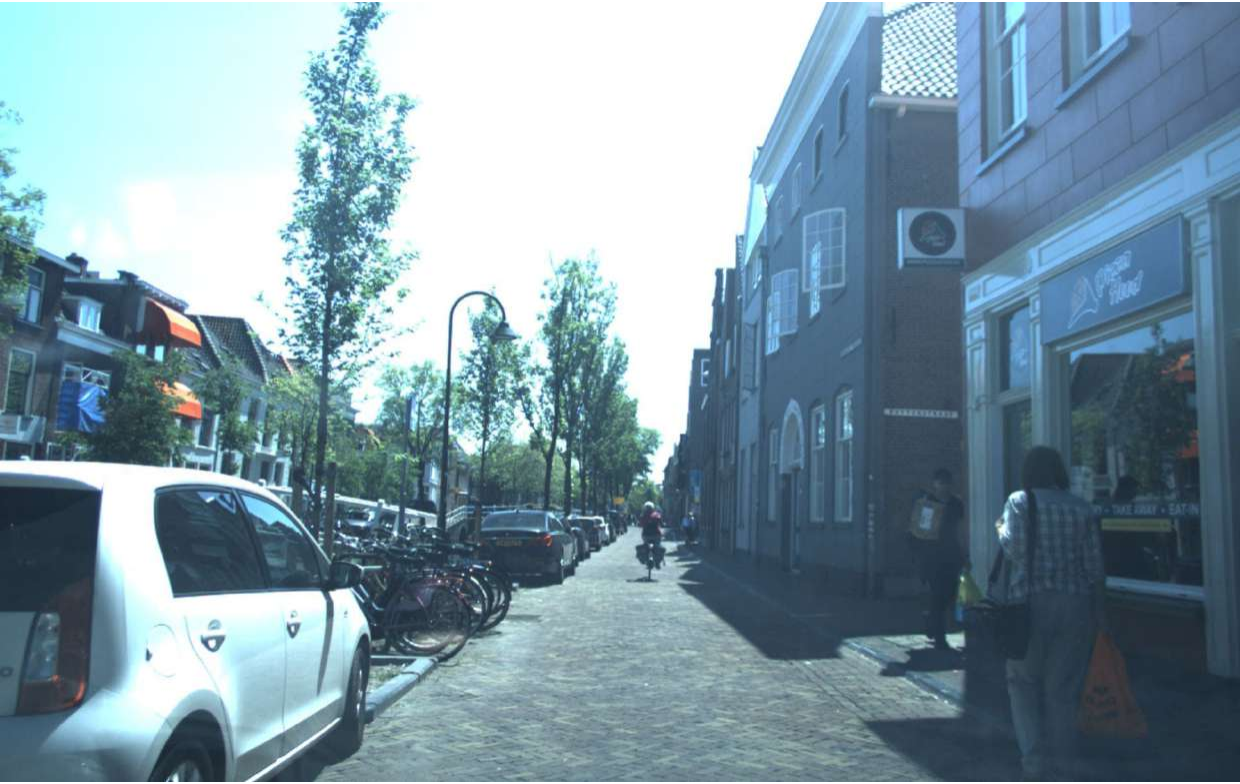}}
  \subfigure[]{\includegraphics[width=0.22\linewidth]{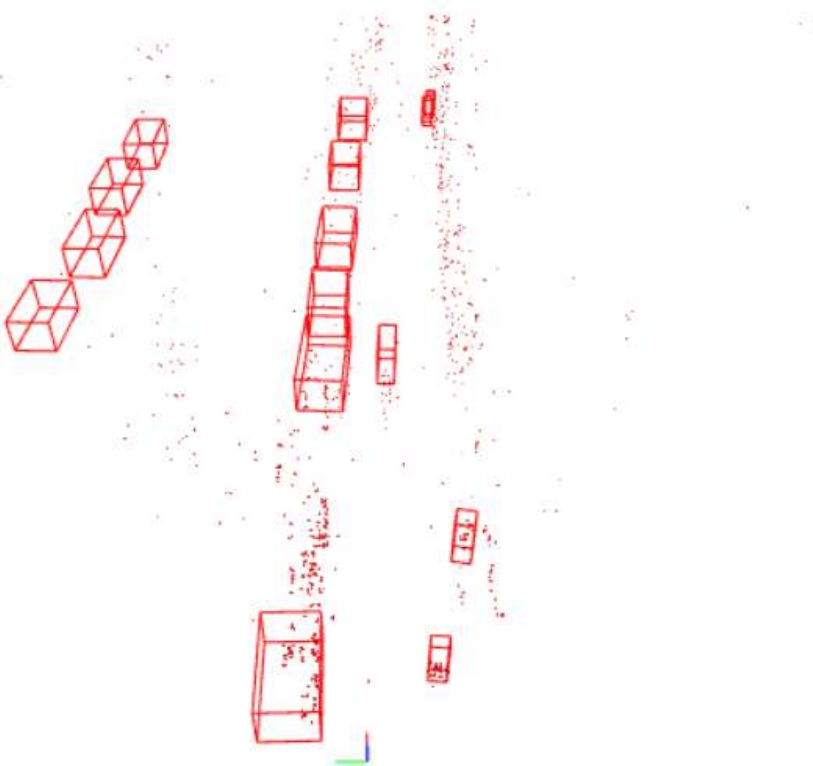}}
  \subfigure[]{\includegraphics[width=0.22\linewidth]{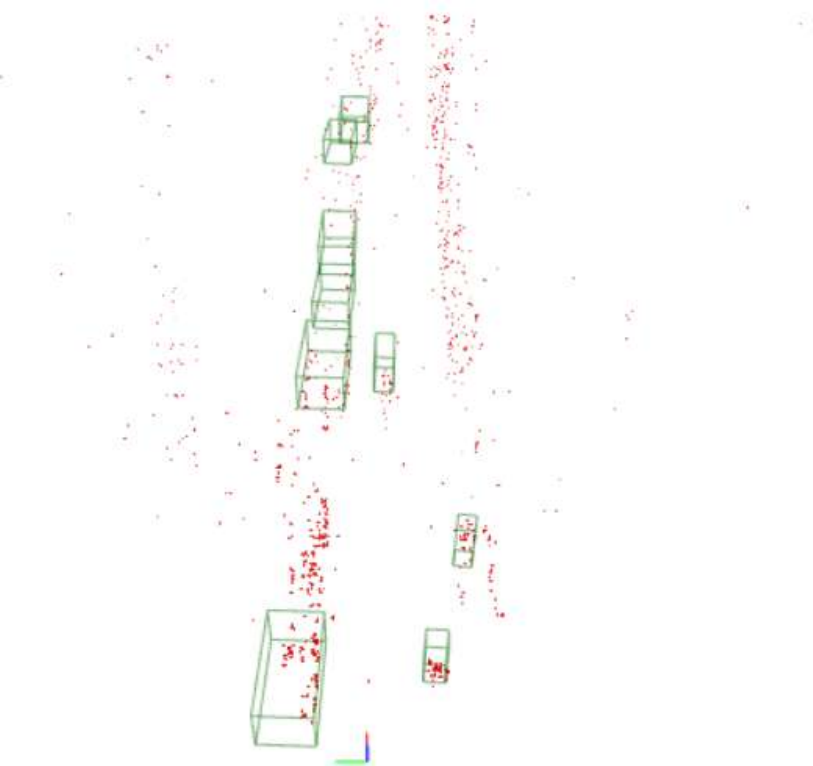}}
  \subfigure[]{\includegraphics[width=0.22\linewidth]{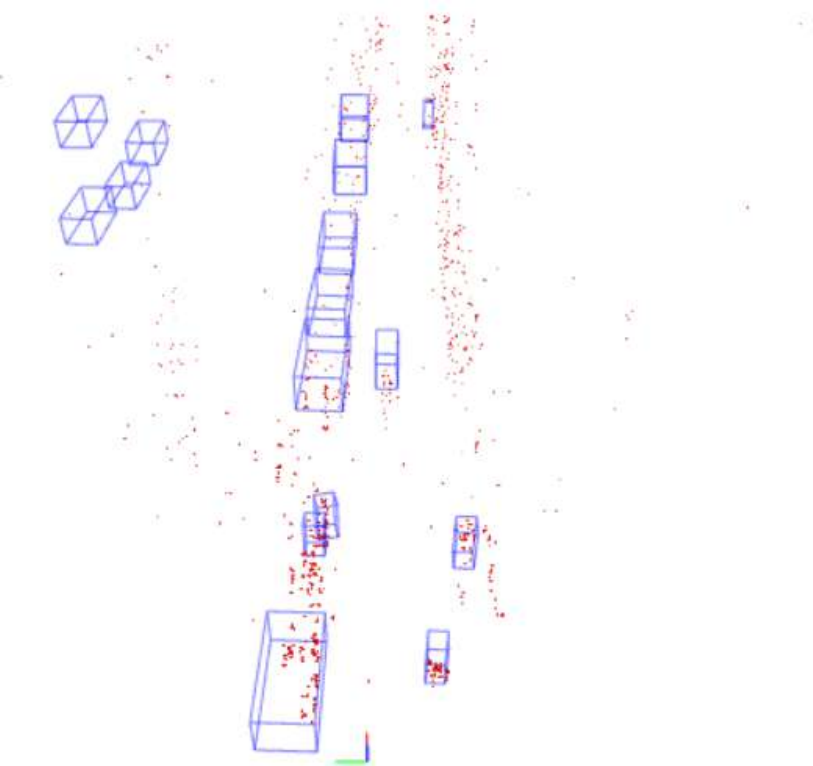}}
  \caption{Some visualizing results on the VoD dataset: (a) represents the image captured by the synchronized cameras. (b) displays the corresponding ground truth 3D bounding boxes for the selected scan, which are highlighted in red. The red dots within the selected scan represent the radar points. (c) shows the predicted bounding boxes obtained from SUMRF without KDE, which are highlighted in dark green. (d) presents the predicted bounding boxes obtained from our proposed SUMRF, which are highlighted in blue.}
  \label{vod_vis}
\end{figure*}

\begin{figure*}[htbp]
  \centering
  \subfigure[]{\includegraphics[width=0.22\linewidth]{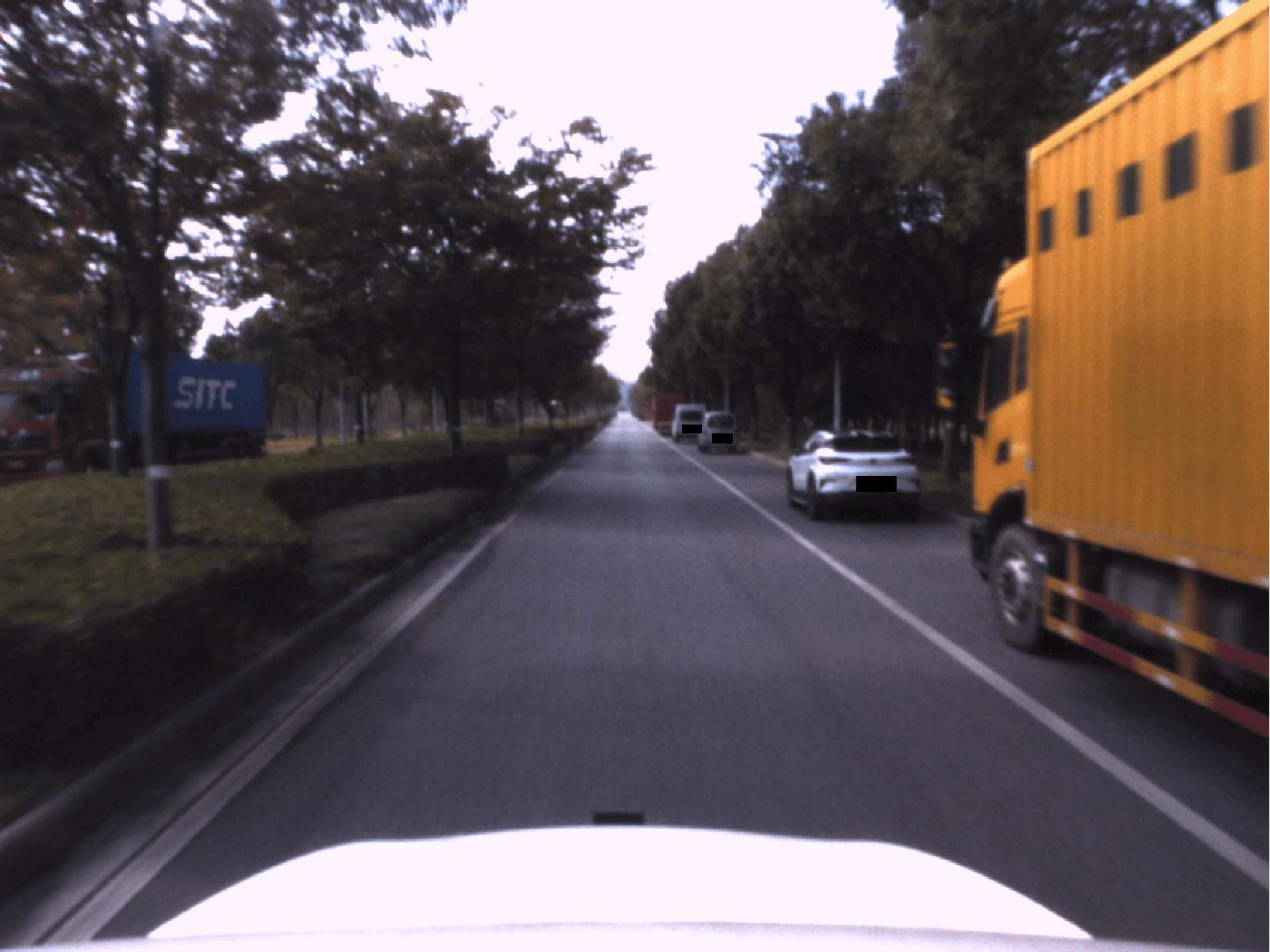}}
  \subfigure[]{\includegraphics[width=0.22\linewidth]{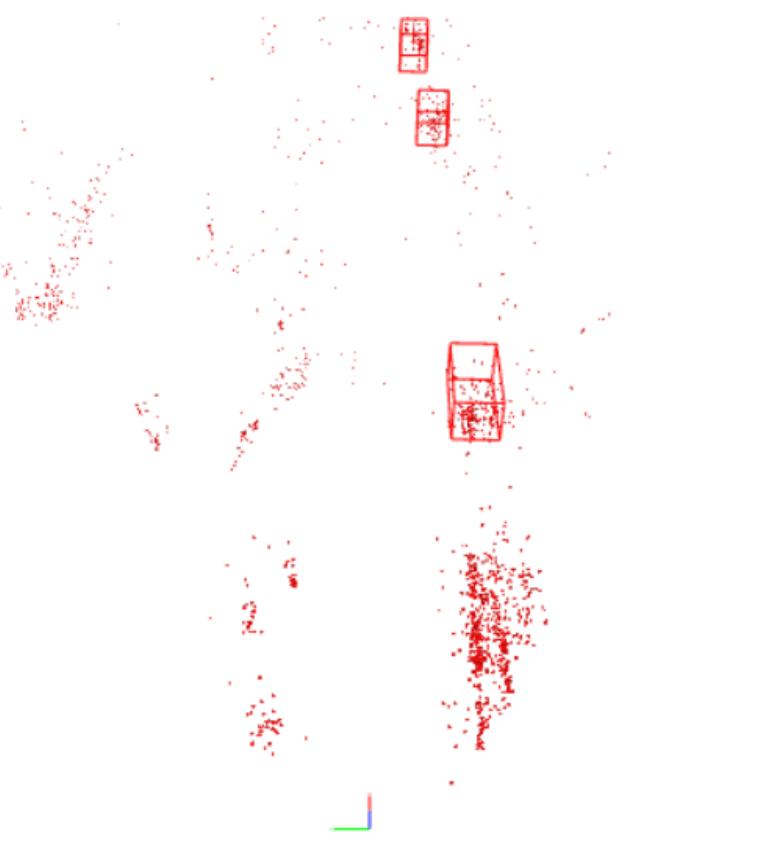}}
  \subfigure[]{\includegraphics[width=0.22\linewidth]{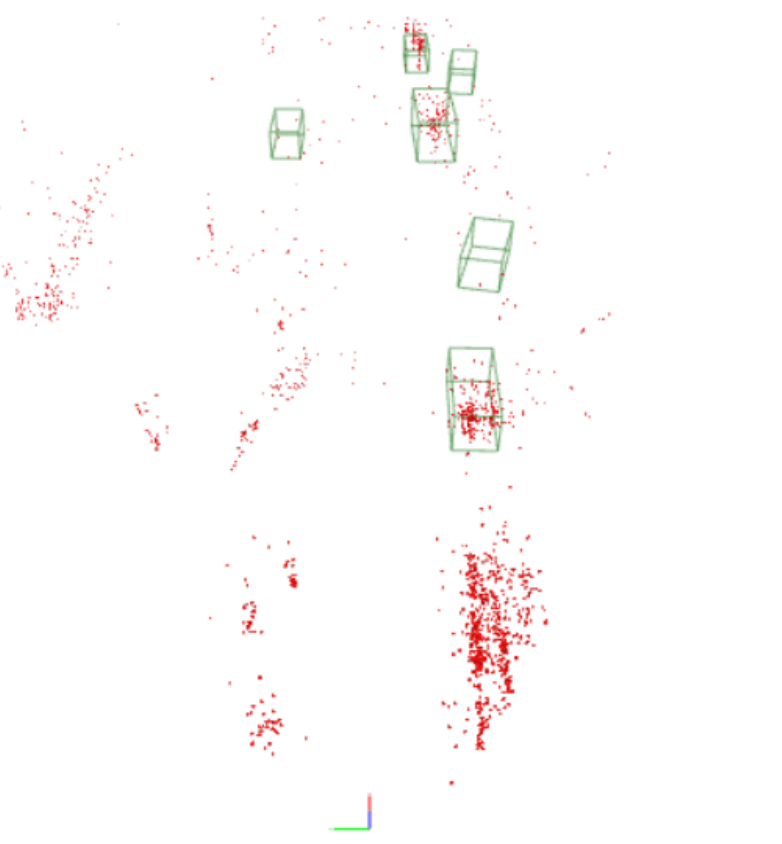}}
  \subfigure[]{\includegraphics[width=0.22\linewidth]{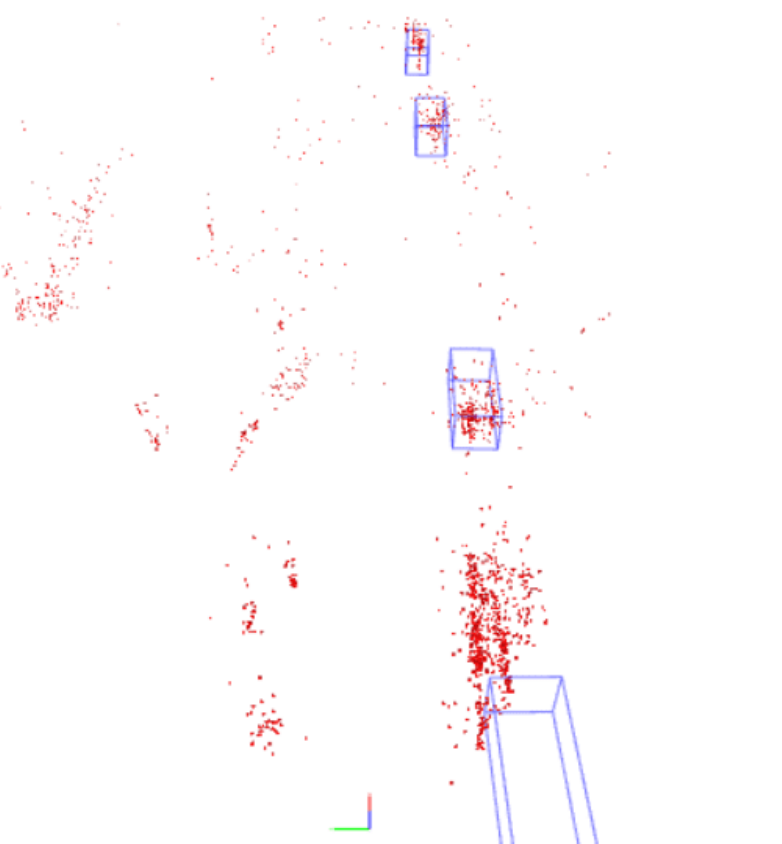}}
  \caption{Visualizing partial results of the TJ4DRadSet. The content and its meaning are congruent with Fig. \ref{vod_vis}}
  \label{tj_vis}
\end{figure*}

To validate the effectiveness of SMURF, ablation experiments were conducted on VoD dataset as a preliminary step. Table \ref{ablation} compares SMURF's performance with two single models: a single KDE feature representation model without the pillarization branch, and a single pillarization representation model without KDE feature extraction.

It can be observed that the model without the pillarization branch demonstrates noticeably poorer performance compared to SMURF. This can be attributed to KDE being introduced solely as an auxiliary density feature extraction tool that doesn't utilize other raw features of the point cloud. Consequently, our subsequent analysis primarily emphasizes the contribution of KDE to enhancing the model's performance rather than focusing on its standalone performance.

The results demonstrate that SMURF achieves improvements across all selected evaluation metrics compared to the one without KDE branch. Specifically, when considering the entire annotated area of the radar, SMURF shows increased $\mathrm{AP_{3D}}$ values for cars, pedestrians, and cyclists. The improvements are observed as a 1.20$\%$ increase for cars, a significant 2.00$\%$ increase for pedestrians, and a notably substantial improvement of 5.58$\%$ for cyclists. This performance enhancement can be attributed to the sparsity of the 4D radar point cloud, which causes smaller objects to have more concentrated points while larger objects may not be fully detected.
Thus, introducing KDE feature extraction contributes to a more significant improvement in the evaluation of smaller objects. Furthermore, the $\mathrm{mAP_{3D}}$ and $\mathrm{mAP_{EBV}}$ values across all three object categories show respective increases of 2.93$\%$ and 1.57$\%$. In the driving corridor, substantial performance improvements are also observed. Specifically, for the $\mathrm{AP_{3D}}$ metric, SMURF achieves increases of 1.73$\%$ for cars, 2.50$\%$ for pedestrians, and 0.69$\%$ for cyclists. Similarly, the $\mathrm{mAP_{3D}}$ and $\mathrm{mAP_{BEV}}$ values across all three categories demonstrate improvements of 1.64$\%$ and 2.76$\%$, respectively.

To further validate the generalization ability of SMURF, an evaluation was conducted on the TJ4DRadSet dataset, and the corresponding performance metrics are also presented in Table \ref{ablation} too. The improvements in $\mathrm{AP_{3D}}$ achieved by SMURF compared to the one without KDE branch are 0.97$\%$ for cars, 1.87$\%$ for pedestrians, a substantial increase of 5.52$\%$ for cyclists, and 3.70$\%$ for trucks. Similarly, when evaluated on the BEV plane, SMURF exhibits superior performance compared to the single-feature model, achieving improvements in $\mathrm{AP_{BEV}}$ for cars, pedestrians, cyclists, and trucks by 4.96$\%$, 2.93$\%$, 6.10$\%$, and 3.88$\%$, respectively. Moreover, the $\mathrm{mAP_{3D}}$ and $\mathrm{mAP_{BEV}}$ values across all four object categories demonstrate increases of 3.02$\%$ and 4.47$\%$, respectively.

Through the conducted ablation experiments on the VoD and TJ4DRadSet datasets, we have established the efficacy of SMURF. By incorporating density features extracted from raw point cloud, our model effectively enriches the information within the point cloud data, leading to improved object detection performance. These findings not only validate the effectiveness of SMURF but also provide a solid basis for further research and experimentation in the field of 4D radar-based object detection.

To visually demonstrate the performance gain brought by our proposed SMURF method, we performed visualizations on a subset of the data, as depicted in Fig. \ref{vod_vis} and Fig. \ref{tj_vis}. The visualizations clearly highlight the advantages of our SMURF method in enhancing the detection capability compared to using a single PointPillars feature representation. Through the introduction of KDE feature representation, SMURF excels in discerning which points are likely to belong to the same object, resulting in a significant improvement in detection accuracy across various scenarios, particularly for small objects such as pedestrians and cyclists, as discussed earlier in this section.

The visual representations show the effectiveness of the SMURF model in reducing missed detections and false alarms for pedestrians and cyclists when compared to a single feature representation model. Furthermore, even for cars with sparsely distributed radar detection points, the SMURF method exhibits a certain degree of detection performance, as exemplified in case (d) in Fig. \ref{vod_vis}.

\subsection{Comparison with Other State-Of-The-Art Methods}
\label{sotacomp}

\begin{table*}[htbp]
  \caption{Comparison on the validation set of VoD dataset. All the approaches are executed with 5-scans radar detection points. The results of PointPillars and RadarPillarNet are inherated from \cite{4d_imaging_radar_camera_3d_object_detection_rcfusion}.}
  \centering
  \begin{tabular}{c|c|c|c|c|c|c|c|c|c|c}
    \hline
    \multirow{2}*{Method} & \multicolumn{5}{c|}{Entire Annotated Area AP} & \multicolumn{5}{c}{In Driving Corridor AP} \\
    \cline{2-11}
    & Car & Ped & Cyc & $\mathrm{mAP_{3D}}$ & $\mathrm{mAP_{BEV}}$ & Car & Ped & Cyc & $\mathrm{mAP_{3D}}$ & $\mathrm{mAP_{BEV}}$ \\
    \hline
    PointPillars (CVPR 2019) \cite{PointPillar} & 37.06 &  35.04 & 63.44 & 45.18 & $N/A$ & 70.15 & 47.22 & 85.07 & 67.48 & $N/A$ \\
    \hline
    CenterPoint (CVPR 2021) \cite{CenterPoint} & 32.74 & 38.00 & 65.51 & 45.42 & 52.75 & 62.01 & 48.18 & 84.98 & 65.06 & 67.05 \\
    \hline
    PillarNeXt (CVPR 2023) \cite{PillarNeXt} & 30.81 & 33.11 & 62.78 & 42.23 & 52.93 & 66.72 & 39.03 & 85.08 & 63.61 & 68.35 \\
    \hline
    RadarPillarNet (IEEE T-IM 2023) \cite{4d_imaging_radar_camera_3d_object_detection_rcfusion} & 39.30 & 35.10 & 63.63 & 46.01 & $N/A$ & 71.65 & 42.80 & 83.14 & 65.86 & $N/A$ \\
    \hline
    SMURF (\textbf{ours}) (x, y, z, Doppler, RCS)
     & \textbf{42.31} & \textbf{39.09} & \textbf{71.50} & \textbf{50.97} & \textbf{56.77} & \textbf{71.74} & \textbf{50.54} & \textbf{86.87} & \textbf{69.72} & \textbf{72.05} \\
    \hline
  \end{tabular}
  \label{vod result}
\end{table*}

\begin{table*}[htbp]
  \caption{Comparison on the test set of TJ4DRadSet dataset. All the approaches are executed with single scan radar detection points. The results of PointPillars, RPFA-Net and RadarPillarNet are inherated from \cite{4d_imaging_radar_camera_3d_object_detection_rcfusion}.}
  \centering
  \begin{tabular}{c|c|c|c|c|c|c|c|c|c|c}
    \hline
    \multirow{2}*{Method} & \multicolumn{5}{c|}{$\mathrm{AP_{3D}}$} & \multicolumn{5}{c}{$\mathrm{AP_{BEV}}$} \\
    \cline{2-11}
    & Car & Ped & Cyc & truck & $\mathrm{mAP_{3D}}$ & Car & Ped & Cyc & truck & $\mathrm{mAP_{BEV}}$ \\
    \hline
    PointPillars (CVPR 2019) \cite{PointPillar} & 21.26 & \textbf{28.33} & 52.47 & 11.18 & 28.31 & 38.34 & \textbf{32.26} & 56.11 & 18.19 & 36.23 \\
    \hline
    CenterPoint (CVPR 2021) \cite{CenterPoint} & 22.03 & 25.02 & 53.32 & 15.92 & 29.07 & 33.03 & 27.87 & 58.74 & 25.09 & 36.18 \\
    \hline
    PillarNeXt (CVPR 2023) \cite{PillarNeXt} & 22.33 & 23.48 & 53.01 & 17.99 & 29.20 & 36.84 & 25.17 & 57.07 & 23.76 & 35.71 \\
    \hline
    RPFA-Net (ITSC 2021) \cite{4d_imaging_radar_3d_object_detection_2} & 26.89 & 27.36 & 50.95 & 14,46 & 29.91 & 42.89 & 29.81 & 57.09 & 25.98 & 38.94 \\
    \hline
    RadarPillarNet (IEEE T-IM 2023 ) \cite{4d_imaging_radar_camera_3d_object_detection_rcfusion} & 28.45 & 26.24 & 51.57 & 15.20 & 30.37 & \textbf{45.72} & 29.19 & 56.89 & 25.17 & 39.24 \\
    \hline
    SMURF (\textbf{ours}) (x, y, z, Doppler, SNR)
     & \textbf{28.47} & 26.22 & \textbf{54.61} & \textbf{22.64} & \textbf{32.99} & 43.13 & 29.19 & \textbf{58.81} & \textbf{32.80} & \textbf{40.98} \\
    \hline
  \end{tabular}
  \label{tj result}
\end{table*}

In this study, we comprehensively evaluate the performance of our SMURF method on the VoD and TJ4DRadSet datasets. The experimental results are summarized in Table \ref{vod result} and Table \ref{tj result}, respectively. To provide a comprehensive comparison, we assess our approach against various 3D object detectors originally designed for LiDAR point cloud, including the anchor-based detector PointPillars, the anchor-free detector CenterPoint, and the state-of-the-art approach PillarNeXt \cite{PillarNeXt}. Additionally, we compare our SMURF method with the recently proposed 3D object detection methods specifically designed for 4D radar, such as RPFA-Net \cite{4d_imaging_radar_3d_object_detection_2} and RadarPillarNet \cite{4d_imaging_radar_camera_3d_object_detection_rcfusion}. To further evaluate the effectiveness of our approach, we compare it with RCFusion \cite{4d_imaging_radar_camera_3d_object_detection_rcfusion}, which represents the latest benchmark in 3D object detection utilizing image and 4D radar fusion. The following paragraphs delve into more detailed experimental results and insights derived from these comparative evaluations.

Our proposed SMURF object detection model demonstrates superior performance compared to several conventional detection methods, showing its high detection accuracy. Evaluating the SMURF method on the VoD dataset reveals remarkable improvements over the anchor-free detector CenterPoint, with an increase of 5.55$\%$ and 4.66$\%$ increase in $\mathrm{mAP_{3D}}$ across the entire annotated area and the driving corridor, respectively. Despite the impressive performance of the anchor-box-free detector PillarNeXt, which achieves state-of-the-art results in 3D object detection with LiDAR point cloud, our SMURF method still outperforms it in terms of $\mathrm{mAP_{3D}}$, with respective improvements of 8.74$\%$ and 6.11$\%$ across the entire annotated area and driving corridor. 
When compared to anchor-box-based detectors like PointPillars and RadarPillarNet, our method consistently demonstrates superior performance, achieving no less than a 4.96$\%$ increase in $\mathrm{mAP_{3D}}$ across the entire annotated area and no less than a 2.24$\%$ increase in $\mathrm{mAP_{3D}}$ in the driving corridor.
Similar to the findings on the VoD dataset, our SMURF method achieves significant performance improvements over many existing methods when evaluated on the TJ4DRadSet dataset as well. However, it is important to note that the detection accuracy of our SMURF method for pedestrians is lower than that of PointPillars, and the $\mathrm{AP_{BEV}}$ for car detection is lower than that of RadarPillarNet. This disparity could be attributed to the scarcity of pedestrian detection points and the near absence of detection points from certain cars in the TJ4DRadSet dataset, which hinder the ability of KDE to effectively extract density features of these objects. Nonetheless, our SMURF method still shows superior performance across various scenarios and object categories, highlighting its effectiveness as a robust 3D object detection framework.

\begin{table*}[htbp]
  \caption{Comparison SMURF which only requires 4D imaging radar, with state-of-the-art 4D imaging radar and camera fusion method on both the VoD validation dataset and TJ4DRadSet test dataset. All the approaches are executed with 5-scans radar detection points on VoD dataset and single-scan radar detection points on TJ4DRadSet dataset for fair comparison.}
  \centering
  \begin{tabular}{c|c|c|c|c|c|c|c|c|c|c}
    \hline
    \multirow{2}*{Method} & \multicolumn{5}{c|}{Entire Annotated Area AP} & \multicolumn{5}{c}{In Driving Corridor AP} \\
    \cline{2-11}
    & Car & Ped & Cyc & $\mathrm{mAP_{3D}}$ & $\mathrm{mAP_{BEV}}$ & Car & Ped & Cyc & $\mathrm{mAP_{3D}}$ & $\mathrm{mAP_{BEV}}$ \\
    \hline
    RCFusion (IEEE T-IM 2023) \cite{4d_imaging_radar_camera_3d_object_detection_rcfusion} & 41.70 & 38.95 & 68.31 & 49.65 & $N/A$ & \textbf{71.87} & 47.50 & \textbf{88.33} & 69.23 & $N/A$ \\
    \hline
    SMURF (\textbf{ours}) (x, y, z, Doppler, RCS)
     & \textbf{42.31} & \textbf{39.09}	& \textbf{71.50}	& \textbf{50.97}	& \textbf{56.77} & 71.74 & \textbf{50.54} & 86.87 & \textbf{69.72} & \textbf{72.05}\\
    \hline
    \hline
    \multirow{2}*{Method} & \multicolumn{5}{c|}{$\mathrm{AP_{3D}}$} & \multicolumn{5}{c}{$\mathrm{AP_{BEV}}$} \\
    \cline{2-11}
    & Car & Ped & Cyc & truck & $\mathrm{mAP_{3D}}$ & Car & Ped & Cyc & truck & $\mathrm{mAP_{BEV}}$ \\
    \hline
    RCFusion (IEEE T-IM 2023) \cite{4d_imaging_radar_camera_3d_object_detection_rcfusion} & \textbf{29.72} & \textbf{27.17} & 
\textbf{54.93} & \textbf{23.56} & \textbf{33.85} & 40.89 & \textbf{30.95}
& 58.30 & 28.92 & 39.76 \\	 
    \hline
    SMURF (\textbf{ours}) (x, y, z, Doppler, SNR)
     & 28.47 & 26.22 & 54.61 & 22.64 & 32.99 & \textbf{43.13} & 29.19 & \textbf{58.81} & \textbf{32.80} & \textbf{40.98} \\
    \hline
  \end{tabular}
  \label{rc result}
\end{table*}

The obtained results highlight the challenges posed by the sparsity of 4D radar point cloud, which results in a limited number of detection points for many objects. Consequently, anchor-free detectors face difficulties in accurately predicting the center of object bounding boxes and other essential information. Additionally, the limitations of the single-pillar representation method in 
extracting discriminative features further hinder the performance of existing approaches. In contrast, our proposed SMURF model addresses these challenges by incorporating a KDE branch. This feature extraction process not only captures valuable density information from the raw point cloud but also mitigates the impact of noise. As a result, SMURF demonstrates superior performance in 3D object detection tasks based on 4D radar.

Despite relying solely on single-modal radar data, SMURF demonstrates competitive performance and sometimes outperforms the latest multi-modal fusion method RCFusion across various metrics.
For instance, SMURF achieves significant improvements in $\mathrm{AP_{3D}}$ for all three object categories across both the entire annotated area and the driving corridor on the VoD dataset, as shown in Table \ref{rc result}. Notably, our method exhibits a remarkable 3.19$\%$ increase in the detection accuracy for bicycles in the entire annotated area. On the TJ4DRadSet dataset, our method demonstrates comparable performance overall and even achieves superior detection results in the BEV domain compared to RCFusion. This highlights the effectiveness of SMURF in leveraging single-modal 4D radar point cloud data for accurate object detection.

\begin{table}[htbp]
  \caption{Inference speed. All results are obtained from the test set of TJ4DRadSet dataset, where the inference speed of RCFusion is sourced from \cite{4d_imaging_radar_camera_3d_object_detection_rcfusion}.}
  \centering
  \begin{tabular}{c|c}
    \hline
    Method & FPS\\
    \hline
    PointPillars (CVPR 2019) \cite{PointPillar} & 42.9\\
    \hline
    CenterPoint (CVPR 2021) \cite{CenterPoint}  & 34.5 \\
    \hline
    PillarNeXt (CVPR 2023) \cite{PillarNeXt} & 28.0\\
    \hline
    RPFA-Net (ITSC 2021) \cite{4d_imaging_radar_3d_object_detection_2} & $N/A$ \\
    \hline
    RadarPillarNet (IEEE T-IM 2023 ) \cite{4d_imaging_radar_camera_3d_object_detection_rcfusion} & $N/A$ \\
    \hline
    RCFusion (IEEE T-IM 2023) \cite{4d_imaging_radar_camera_3d_object_detection_rcfusion}& 4.7$-$10.8 \\
    \hline
    SMURF (\textbf{ours}) (x, y, z, Doppler, SNR) & 23.1 \\
    \hline
  \end{tabular}
  \label{speed result}
\end{table}

Furthermore, the utilization of image data typically requires substantial computational resources due to the large volume of data. In contrast, SMURF achieves remarkable performance solely with 4D radar point cloud data, while using fewer computational resources. Specifically, our SMURF model implemented in the MMDetection3D framework \cite{mmdetection3d}, exhibits efficient inference times. On the validation set of the VoD dataset, the average inference time per scan is approximately 0.033 seconds, while on the test set of the TJ4DRadSet dataset, it is around 0.043 seconds.
In other words, the inference speed of SMURF exhibits a lower bound of 23 FPS, which outperforms the range of 4.7$-$10.8 FPS achieved by model RCFusion as shown in Table \ref{speed result}. It is sufficient to consider a frame rate of at least 10 FPS for real-time object detection applications \cite{real-time 2}. Despite slightly slower inference speed of SMURF compared to  PillarNeXt, as shown in Table \ref{speed result}, a frame rate of 23 FPS is adequate to fulfill the real-time detection requirements. Besides, our inference speed is on par with works claiming real-time detection, like \cite{real-time 1}\cite{real-time 2}\cite{real-time 3} with normal inference speeds ranging from 10 to 30 FPS, further indicating that SMURF effectively meets the requirements for real-time detection. These results demonstrate that SMURF can effectively meet the requirements for real-time object detection.

%===============================================
\section{Conclusion}
\label{conclusion}

The utilization of 4D radar data poses challenges due to the sparsity of detection points, noises, and the limited feature extraction capacity of existing methods. In this paper, we propose the SMURF method for 3D object detection using 4D imaging radar point cloud. 
In the feature encoding stage, we adopt a pillar-based point cloud representation to minimize computational overhead. At the same time, to mitigate the adverse effect caused by inherent noise in the original point cloud and extract richer semantic information from sparse point cloud, we introduce the KDE method. By fusing multiple-representation features of point cloud, we successfully enhance the precision of 3D object detection.

Through extensive experimental evaluation on VoD and TJ4DRadSet datasets, we have demonstrated the effectiveness and generalization capability of SMURF. Our results indicate that SMURF offers competitive performance when compared to the state-of-the-art multi-modal model that leverages both 4D radar and images information. However, the introduction of KDE does not effectively eliminate the impact of background objects, as points from background objects such as trees can also be dense. We leave this problem for future work.

By showing superior performance, SMURF has the potential to serve as a robust baseline for future research in the field of 3D object detection using 4D radar. Its compact architecture and high inference speed further enhance its viability for practical applications that require real-time object detection capabilities. 
The improved detection accuracy positions SMURF as a promising solution for real-world scenarios, and could be potentially helpful for subsequent downstream tasks like planning and control \cite{wang2023adaptive}\cite{zheng2023control}.

\end{document}